\pdfoutput=1

\documentclass[11pt,a4paper]{article}
\usepackage{times,latexsym}
\usepackage{url}
\usepackage[T1]{fontenc}
\usepackage[acceptedWithA]{tacl2021v1}
\usepackage{tacl2021v1}
\usepackage{microtype}
\usepackage{inconsolata}
\usepackage{tacl2021v1}
\usepackage{booktabs}
\usepackage{multirow}
\usepackage{makecell}
\usepackage{amsmath}
\usepackage{setspace}
\usepackage{graphicx}
\usepackage{xspace}
\usepackage{CJKutf8}
\usepackage[utf8]{inputenc}
\usepackage{tabularx}
\usepackage{hyperref} 
\usepackage{ruby}
\usepackage{xpinyin}
\usepackage{ulem}

\newcommand{\cy}[1]{#1}
\newcommand{\bkcolorwrong}[1]{\textcolor[HTML]{F02A2A}{#1}}
\newcommand{\bkcolorliteral}[1]{\textcolor[HTML]{999999}{#1}}
\newcommand{\bkcolordiff}[1]{\textcolor[rgb]{0.75, 0.5, 0.25}{#1}}
\newcommand{\bkcoloringreunit}[1]{\textcolor[HTML]{08b1f0}{#1}}
\newcommand{\bkcoloringre}[1]{\textcolor[HTML]{548236}{#1}}
\newcommand{\bkcolortool}[1]{\textcolor[HTML]{7438a3}{#1}}
\newcommand{\bkcoloraction}[1]{\textcolor[HTML]{ff9300}{#1}}
\makeatletter
  \UL@protected\def\reduwave{\leavevmode \bgroup 
    \ifdim \ULdepth=\maxdimen \ULdepth 3.5\p@
    \else \advance\ULdepth2\p@ 
    \fi \markoverwith{\lower\ULdepth\hbox{\textcolor{red}{\sixly \char58}}}\ULon}
\makeatother

\newenvironment{myquote}[1]%
  {\list{\vartriangleright}{\leftmargin=#1\rightmargin=#1}\item[]}%
  {\endlist}

\title{Cultural Adaptation of Recipes}

\newcommand{\hust}{$^1$}
\newcommand{\ku}{$^2$}
\newcommand{\uestc}{$^3$}
\newcommand{\saarland}{$^4$}
\newcommand{\eqcontrib}{$^*$}

\author{Yong Cao\hust$^,$\ku\eqcontrib, Yova Kementchedjhieva\ku\eqcontrib, Ruixiang Cui\ku, Antonia Karamolegkou\ku, \\ 
{\bf Li Zhou\ku$^,$\uestc, Megan Dare\saarland, Lucia Donatelli\saarland and Daniel Hershcovich\ku} \\
  {\hust}Huazhong University of Science and Technology \\
  {\ku}Department of Computer Science, University of Copenhagen \\
  {\uestc}University of Electronic Science and Technology of China \\
  {\saarland}Department of Language Science and Technology, Saarland University \\
\texttt{\{yongcao,yova,rc,antka,dh\}@di.ku.dk,li\_zhou@std.uestc.edu.cn, } \\ \texttt{\{mdare,donatelli\}@coli.uni-saarland.de}}

\begin{document}
\maketitle
\def\thefootnote{*}\footnotetext{Equal contribution.}\def\thefootnote{\arabic{footnote}}
\begin{abstract}
Building upon the considerable advances in Large Language Models (LLMs), we are now equipped to address more sophisticated tasks demanding a nuanced understanding of cross-cultural contexts. A key example is recipe adaptation, which goes beyond simple translation to include a grasp of ingredients, culinary techniques, and dietary preferences specific to a given culture. We introduce a new task involving the translation and cultural adaptation of recipes between Chinese and English-speaking cuisines. To support this investigation, we present CulturalRecipes, a unique dataset comprised of automatically paired recipes written in Mandarin Chinese and English. This dataset is further enriched with a human-written and curated test set. In this intricate task of cross-cultural recipe adaptation, we evaluate the performance of various methods, including GPT-4 and other LLMs, traditional machine translation, and information retrieval techniques. Our comprehensive analysis includes both automatic and human evaluation metrics. While GPT-4 exhibits impressive abilities in adapting Chinese recipes into English, it still lags behind human expertise when translating English recipes into Chinese. This underscores the multifaceted nature of cultural adaptations. We anticipate that these insights will significantly contribute to future research on culturally-aware language models and their practical application in culturally diverse contexts.
\end{abstract}

\section{Introduction}\label{sec:introduction}

Cooking recipes are a distinct form of procedural text whose accurate interpretation depends on several factors. Familiarity with ingredients and measurement units, common sense about the cooking environment, and reasoning about how tools and actions affect intermediate products in the cooking process are necessary to successfully craft a recipe. Such knowledge varies by culture and language, 
as a result of geography, history, climate, and economy \cite{albala2012three}. These factors impact the frequency of ingredients usage, the available forms and cost of heat for cooking, common taste profiles, written recipe style, etc. (\S\ref{sec:differences}). 

Identifying and adapting to cultural differences in language use is important and challenging \cite{hershcovich-etal-2022-challenges}.
Recipe translations with current machine translation technology may gloss over culture-specific phraseology or yield 
mistranslations due to a lack of grounding in the physical and cultural space. Literal translations are often opaque or odd: a Chinese dish, \begin{CJK}{UTF8}{gbsn}\xpinyin*{夫妻肺片}\end{CJK} (literally, `husband and wife lung slices'), can be adapted in translation to `Sliced Beef in Chili Sauce' for English-speaking cooks.
Structural patterns in recipes in different cultures (e.g., \textit{mise en place}\footnote{In French cooking, \textit{mise en place} is the practice of measuring out and cutting all ingredients in advance. }) additionally make straightforward recipe translation difficult: cuisines differ in dish preparation methods, and temporal dependencies between actions complicate the disentanglement of recipe actions \cite{kiddon2015mise,yamakata2017comparison}.

In this work, we introduce the task of adapting cooking recipes across languages and cultures. Beyond direct translation, this requires adaptation with respect to style, ingredients, measurement units, tools, techniques, and action order preferences.
Focusing on recipes in Chinese and English, we automatically match pairs of recipes for the same dish drawn from two monolingual corpora, and train text generation models on these pairs. We evaluate our methodology with human judgments and a suite of automatic evaluations on a gold standard test set that we construct. We provide ample evidence that recipe adaptation amounts to more than mere translation and find that models finetuned on our dataset can generate grammatical, correct, and faithful recipes, appropriately adapted across cultures.
\cy{Intriguingly, Large Language Models (LLMs) outperform our finetuned models in both automatic and human evaluations, even without training on our paired dataset. This unexpected result opens multiple avenues for future research, including how large-scale pre-training could complement our dataset and nuanced evaluation metrics that could better capture the complexities of recipe adaptation.}
Our contributions are as follows:

(a) We introduce the task of cross-cultural recipe adaptation and build a bidirectional Chinese-English dataset for it, \textbf{CulturalRecipes} (\S\ref{sec:dataset}).

(b) We experiment with various sequence-to-sequence approaches to adapt the recipes, including machine translation models and multilingual language models (\S\ref{sec:experiments}).

(c) We evaluate and analyze the differences between Chinese and English-speaking cultures as reflected in the subcorpora (\S\ref{sec:alignment}) and to the translation and adaptation of recipes (\S\ref{sec:experiments}).

Our dataset, code, and trained models will be freely available upon publication. 


\begin{figure}[t]
	\centering
	\includegraphics[width=1.0\columnwidth]{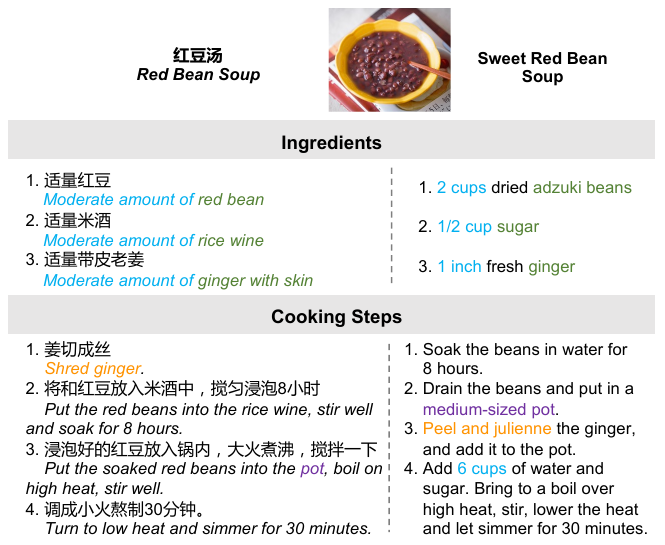}
	\caption{An example of cultural differences between Chinese (left) and English (right) recipes by color: \bkcoloringreunit{blue} text signals contrasts in ingredient measurement units; \bkcoloringre{green}, ingredients; \bkcoloraction{orange}, actions performed by cooks; and \bkcolortool{purple}, tools. For readability, we show our literal translation on the left along with the original Chinese.}
	\label{fig:figure1}
\end{figure}

\section{Cultural Differences in Recipes}
\label{sec:differences}

Extensive cross-cultural culinary research reveals compelling differences in ingredients, measurement units, tools, and actions, each reflecting historical, geographical, and economic influences unique to each culture \cite{albala2012three}. For example, the historical reliance on open flame cooking in China has cultivated an array of oil-based cooking techniques exclusive to Chinese cuisine. Further complexities arise from culture-specific terminologies for cooking methods and dish names, which pose formidable challenges to translation and adaptation \cite{10.1007/978-3-319-69805-2_8}. Additionally, the visual presentation of online recipes exhibits striking contrasts across different cultural contexts \cite{zhang2019understanding}. Delving deeper, culinary preferences also demonstrate regional patterns in flavor profiles; Western cuisines tend to combine ingredients that share numerous flavor compounds, while East Asian cuisines often intentionally avoid such shared compounds \cite{ahn2011flavor}. These intricate cultural nuances underscore the complexity and diversity inherent in global culinary practices, thereby emphasizing the intricacy involved in adapting recipes across different cultures.

\paragraph{Examples.} Figure~\ref{fig:figure1} presents \cy{a Mandarin Chinese recipe and its human-authored adaptation to American English, highlighting key differences:}

\noindent(1) \textit{Ingredients.} Distinct ingredients feature prominently in each recipe; the Chinese version highlights \begin{CJK}{UTF8}{gbsn}\xpinyin*{米酒}\end{CJK} `rice wine', \begin{CJK}{UTF8}{gbsn}\xpinyin*{红豆}\end{CJK} `red beans', and \begin{CJK}{UTF8}{gbsn}\xpinyin*{带皮老姜}\end{CJK} `ginger with skin'. Interestingly, while `red bean' is referenced in Chinese recipes, the equivalent ingredient is typically recognized as `adzuki beans' in Western countries.

\noindent(2) \textit{Measurement units.} Chinese recipes often rely on imprecise measurements, guided by the cook's experience, while American English recipes use precise U.S. customary or Imperial units like `cups', `inches', `pints', and `quarts'. Occasionally, Chinese recipes employ traditional units such as \begin{CJK}{UTF8}{gbsn}\xpinyin*{两}\end{CJK} and \begin{CJK}{UTF8}{gbsn}\xpinyin*{斤}\end{CJK}, or metric system units like `grams (g)' and `milliliters (ml)'.

\noindent(3) \textit{Tools.} Specificity varies between recipes, with English recipes typically specifying pot sizes while Chinese recipes provide more general descriptions. Chinese recipes also favor stovetop cooking over ovens, contrasting with their English counterparts.

\noindent(4) \textit{Actions by cook.} Preparation methods often vary between Chinese and English recipes. For instance, Chinese recipes usually involve shredding ginger, while English recipes recommend peeling and julienning. Additionally, unique processes like \begin{CJK}{UTF8}{gbsn}\xpinyin*{焯水}\end{CJK} 'blanching', common in Chinese cooking to remove unwanted flavors, are rarely found in English recipes. These differences highlight the subtle cultural nuances in similar recipes.

\paragraph{Over-generalization and bias.}
\label{sec:bias_limit}
In a study of cultural adaptation, it is important to recognize that the concept of ``culture'' is multifaceted and complex. When we refer to Chinese- and English-speaking cultures throughout this work, we make the simplifying assumption that there are general features that characterize the cooking of these cultures and make them distinct in certain systematic ways. We recognize that there is enormous diversity within these simplistic categories,\footnote{For example, southern and northern Chinese cuisines are vastly different, with rice  and wheat as staples respectively.} but as a first step towards the adaptation of recipes across cultures, we restrict ourselves to the coarse-grained level only.  

To enable the development and benchmarking of recipe adaptation, we build a dataset for the task.

\section{The CulturalRecipes Dataset}\label{sec:dataset}
Our dataset, \textit{CulturalRecipes}, builds on two existing large-scale recipe corpora in English and Chinese, respectively.
We create two collections of automatically paired recipes, one for each direction of adaptation (English$\to$Chinese and Chinese$\to$English), which we use for training and validation in our recipe adaptation experiments (\S\ref{sec:experiments}).
Additionally, \textit{CulturalRecipes} incorporates a small test set of human adaptations expressly crafted for the task in each direction, serving as references in our experimental evaluation.

\subsection{Recipe Corpora}\label{sec:sources}
We source recipes from two monolingual corpora:
\textbf{RecipeNLG} \cite{bien-etal-2020-recipenlg} and \textbf{XiaChuFang} \cite{liu-etal-2022-counterfactual}.\footnote{For license details, please refer to \url{https://recipenlg.cs.put.poznan.pl/dataset} for RecipeNLG and \url{https://xiachufang.com/principle} for XiaChuFang}
RecipeNLG consists of over 2M English cooking recipes. It is an extension of \textsc{Recipe1M}~\cite{salvador2017learning} and \textsc{Recipe1M+}~\cite{marin2019learning}, with improvements in data quality.
XiaChuFang consists of 1.5M recipes from the Chinese recipe website \url{xiachufang.com}, split into a training and evaluation set. We use the training set and clean it by removing emojis,\footnote{Despite their potential significance, we remove emojis since they occur only in a few XiaChuFang recipes.} special symbols, and empty fields.   We use the title, ingredients, and cooking steps fields of the recipes from both corpora.
The recipes in RecipeNLG consist of nine ingredients and seven steps on average, and in XiaChuFang, of seven ingredients and seven steps.
As these two corpora are independent and monolingual, discovering recipe equivalents between them is not trivial.

\begin{figure}
    \centering
    \includegraphics[width=\linewidth]{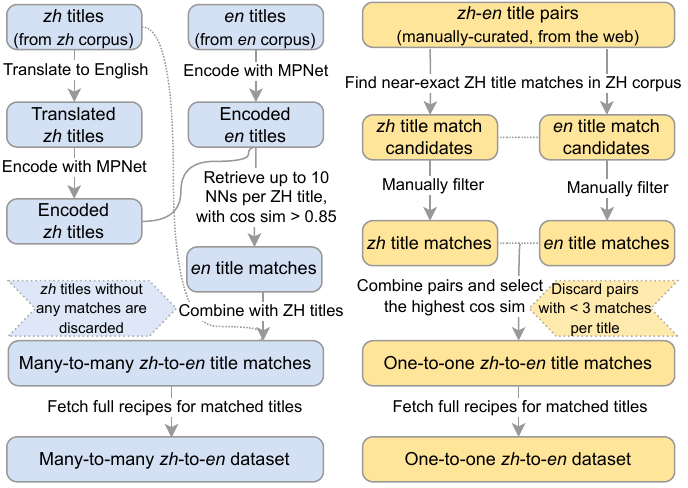}
    \caption{Training and validation (left) and test (right) silver-standard data compilation in the direction Chinese$\to$English. The process is analogous for the opposite direction.}
    \label{fig:flowchart}
\end{figure}

\subsection{Recipe Matching Rationale}
Our recipe matching procedure relies on the following assumption: if two recipes have the same title, they describe the same dish. This assumption can be applied even in a monolingual context: if two recipes are both titled `Veggie Lasagna', we can assume that they describe the same dish \cite{lin-etal-2020-recipe,donatelli-etal-2021-aligning}. It is permissible that there is some mismatch in the set of ingredients, in the number and sequence of steps, in the measurement units and exact amounts, etc. The same assumption can be said to hold for a recipe with a slightly different, but semantically equivalent title, e.g., `Vegetable Lasagna'. Similarly, if we take the Chinese recipe title \begin{CJK}{UTF8}{gbsn}\xpinyin*{卷心菜番\xpinyin{茄}{qie2}牛肉汤}\end{CJK}, we translate it to `Cabbage tomato beef soup' and we find a recipe with a very similar title in English, e.g., `Cabbage beef soup', we can assume that these two recipes describe the same dish. The degree to which this assumption holds depends on the quality of translation of recipe titles from one language into the other, on the measure of similarity, and on how much distance we allow for between two recipe titles before they are no longer considered semantically equivalent. These factors guide our approach to building a silver-standard dataset for the task, further described below, with the procedure also visualized in Figure~\ref{fig:flowchart}, and the statistics of the resulting datasets reported in Table~\ref{tab:dataset_info}.\footnote{Prior to the procedure described below, we filter out recipes longer than 512 subword tokens \cite[arbitrarily using the mT5 tokenizer;][]{xue-etal-2021-mt5} to facilitate using the neural approaches described in \S\ref{sec:experiments}.}

\subsection{Silver-standard Data}\label{sec:silver}

\paragraph{Training and validation sets.}


\begin{figure}
    \centering
    \includegraphics[width=\columnwidth]{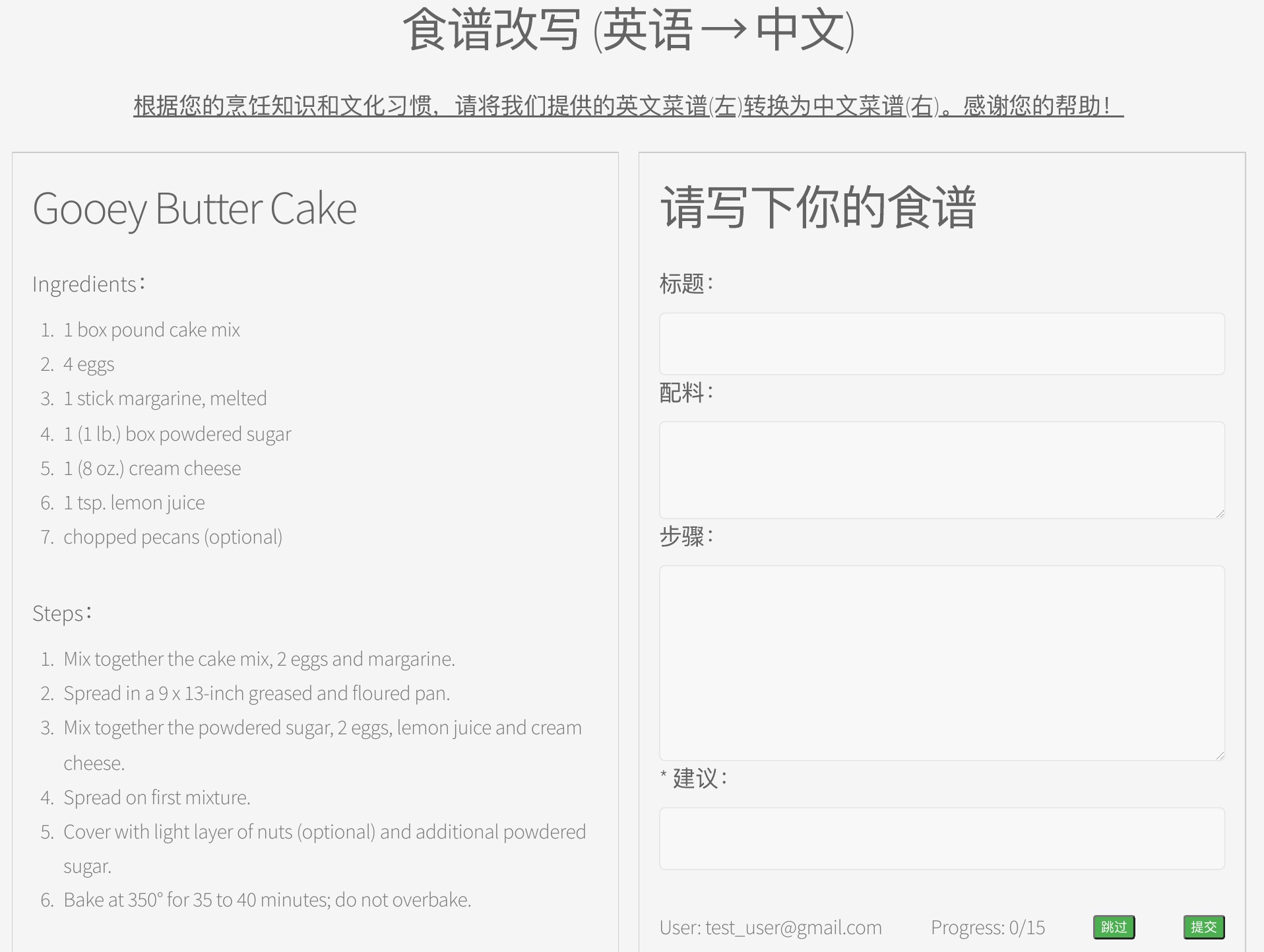}
    \caption{Screenshot from our human recipe adaptation platform, demonstrating the English$\to$Chinese direction, with the source recipe on the left. On the right, participants should adapt the title, ingredients and steps based on their culinary knowledge and cultural habits.}
    \label{fig:zh-en-adaptation-screenshot}
\end{figure}

We obtain training recipe pairs by (1) automatically translating all recipe titles in the Chinese corpus to English using a pre-trained machine translation model \cite{TiedemannThottingal:EAMT2020};\footnote{\href{https://huggingface.co/Helsinki-NLP/opus-mt-zh-en}{\texttt{Helsinki-NLP/opus-mt-zh-en}}} (2) encoding all English and translated Chinese titles with the MPNet sentence encoder \cite{NEURIPS2020_c3a690be}\footnote{\href{https://huggingface.co/sentence-transformers/all-mpnet-base-v2}{\texttt{sentence-transformers/all-mpnet-base-v2}}} to obtain two embedding spaces; and (3) in each direction (English$\to$Chinese and Chinese$\to$English), retrieving up to $k=10$ nearest neighbors per source title from the target space, and filtering out any neighbors that have a cosine similarity against the source title lower than 0.85.\footnote{The similarity threshold for retrieval was chosen through manual inspection of the quality of retrieved pairs.}
The resulting sets, one in each direction, contain multiple reference targets for each source recipe.
We further split the matches into training and validation sets.

We recognize that the aforementioned procedure can be susceptible to various sources of noise due to the translation of titles, the encoder representations, and the fixed similarity threshold. We trust that the signal-to-noise ratio should still be sufficient to enable model learning, but for evaluation we need cleaner, more representative data.

\paragraph{Test set.}
We are able to eliminate one of the aforementioned sources of noise by collecting manual translations of Chinese recipe titles into English and vice versa from websites that explicitly mention the original dish name when presenting an adapted version.\footnote{For Chinese$\to$English we use \href{https://omnivorescookbook.com/recipe-filter/}{\textit{Easy Chinese Recipes}}, \href{https://thewoksoflife.com/category/recipes/page/0/}{\textit{Recipes Archives}}, \href{https://healthynibblesandbits.com/category/asian-food/}{\textit{Asian Food Archives}}, \href{https://redhousespice.com/index/}{\textit{Authentic Chinese Recipes}}; for English$\to$Chinese, \href{https://www.christinesrecipes.com/p/western-recipes.html}{\textit{Christine’s Recipes}} and \href{https://www.wikipedia.org/}{Wikipedia}. We convert any traditional Chinese text to simplified Chinese using \href{https://pypi.org/project/zhconv/}{\texttt{zhconv}} to match our other data sources.} This should resolve issues like \begin{CJK}{UTF8}{gbsn}\xpinyin*{夫妻肺片}\end{CJK} being translated literally by an automatic MT system (see \S\ref{sec:introduction}). To supplement these titles with a corresponding list of ingredients and steps, we look up each title in the recipe corpus of the corresponding language and find the most similar title within, allowing for different capitalization, punctuation and slight differences in word choice and order, e.g., `Rice with caramelized leeks' and `Caramelized Leek Rice' (we manually inspect candidate matches to ensure semantic equivalence).

The resulting test set closely resembles the training data, thus allowing us to determine how well the models we train do in the setting they were trained for (mapping between automatically matched recipes). In order to evaluate the models' ability to perform the true task we want to solve, i.e. adapting specific recipes from one culture to another, we also construct a gold-standard test set.

\begin{table}[t]
	\resizebox{\columnwidth}{!}{
		\begin{tabular}{ccl|cc|cc}
			\toprule
                &&& \multicolumn{2}{c|}{\textbf{\# Recipes}} & \multicolumn{2}{c}{\textbf{Mean \# Tokens}}\\
                &&& \small\textbf{Source} &  \small\textbf{Target} & \small\textbf{Source} &  \small\textbf{Target}\\
			\midrule
			  \multirow{2}{*}{\rotatebox[origin=c]{90}{\footnotesize{\textbf{Train}}}} & \hspace{-0.5cm} \multirow{2}{*}{\rotatebox[origin=c]{90}{\footnotesize{\textbf{\& Val}}}} & \textit{zh}$\to$\textit{en} & 44,5k & 144,6k & 159.1  &  140.2 \\
			&&\textit{en}$\to$\textit{zh} & 43,8k  & 120,7k  & 117.1 & 164.8\\
			\midrule
                \multirow{2}{*}{\rotatebox[origin=c]{90}{\footnotesize{\textbf{Silver}}}} & \hspace{-0.5cm} \multirow{2}{*}{\rotatebox[origin=c]{90}{\footnotesize{\textbf{Test}}}} & \textit{zh}$\to$\textit{en} & 82 & 82 & 140.5 & 144.7 \\
			&&\textit{en}$\to$\textit{zh} & 52 & 52 & 122.7 & 153.3 \\
			\midrule
                \multirow{2}{*}{\rotatebox[origin=c]{90}{\footnotesize{\textbf{Gold}}}} &\hspace{-0.5cm} \multirow{2}{*}{\rotatebox[origin=c]{90}{\footnotesize{\textbf{Test}}}} & \textit{zh}$\to$\textit{en} & 25 & 25 & 139.8 & 97.1 \\
			&&\textit{en}$\to$\textit{zh} & 41 & 41 & 115.7 & 176.5 \\
			\bottomrule
	\end{tabular}}
	\caption{\label{tab:dataset_info} Statistics of (many-to-many) training, (one-to-one) silver-standard and gold-standard (human-written) evaluation sets for both directions. \textit{zh}: Chinese. \textit{en}: English. 
    We count tokens with whitespace tokenization for English and \href{https://github.com/fxsjy/jieba}{\texttt{jieba}} text segmentation for Chinese.}
\end{table}

\subsection{Gold-standard Test Data}\label{sec:humanadap}%
We include human-written adaptations in our dataset as the ground truth for reference-based evaluations (\S\ref{sec:autoeval}, \S\ref{sec:structeval}) and as a point of comparison in human evaluations (\S\ref{sec:humaneval}). We select 41 English recipes and 25 Chinese recipes manually from the silver test sets to adapt each to the other culture.

We develop an in-house web application as our recipe writing platform, illustrated in Figure~\ref{fig:zh-en-adaptation-screenshot}. Our guidelines encourage participants to adapt recipes based on their culinary knowledge and cultural customs. We give participants the option to skip a recipe if they are not able to confidently adapt it. Six native Chinese speakers proficient in English with experience in both Chinese and Western cooking volunteered for the task, spending 6.4 minutes on average to adapt a recipe. 
Subsequently, three of the authors, fluent in both English and Chinese, who have substantial cooking experience, hand-corrected and improved all adapted recipes, including filtering incomplete source recipes, and correcting grammatical errors, spelling mistakes, and non-executable recipe expressions.

\section{Corpus Analysis}\label{sec:alignment}  
Here, we perform a data-driven analysis to investigate how the cultural differences discussed in \S\ref{sec:differences} are realized in English and Chinese recipe corpora through the lens of distributional semantics.

\subsection{Embedding Alignment}
In this analysis, we train static monolingual word embeddings on English and Chinese recipe data, respectively, as a means of capturing their distributional properties.
While the global geometry of English and Chinese distributional spaces is similar \cite{lample2018word}, we hypothesize that cultural differences would lead to mismatches in the local geometry of the two spaces \cite{sogaard-etal-2018-limitations}. We test this hypothesis through cross-lingual embedding alignment, wherein the English and Chinese embeddings are aligned through a linear mapping to obtain a cross-lingual embedding space, in which semantic equivalents between the two languages should occupy a similar position. 

We train monolingual word embeddings using Word2Vec based on a skipgram model by \cite{Mikolovw2v} on the entire English and Chinese corpora (\S\ref{sec:silver}),\footnote{We train 300-dimensional embeddings for 5 epochs using a minimum frequency count of 10, window size of 5, and 10 negative samples. Chinese text is tokenized with \href{https://github.com/fxsjy/jieba}{\texttt{jieba}}.} and align them using VecMap \cite{artetxe2017acl} 
with weak supervision from a seed dictionary of 15 culturally neutral word pairs we manually curate.\footnote{Seed dictionary: \begin{CJK}{UTF8}{gbsn}spinach-\xpinyin*{菠菜}, onion-\xpinyin*{洋葱}, flour-\xpinyin*{面粉}, potatoes-\xpinyin*{土豆}, egg-\xpinyin*{蛋}, salt-\xpinyin*{盐}, sugar-\xpinyin*{糖}, apples-\xpinyin*{苹果}, mix-\xpinyin*{混合}, chop-\xpinyin*{劈}, pour-\xpinyin*{倒}, knife-\xpinyin*{刀}, bowl-\xpinyin*{碗}, pot-\xpinyin*{锅}, chicken-\xpinyin*{鸡}.\end{CJK}} 

\subsection{Analysis}
We use the top 100 most common Chinese content words in the XiaChuFang dataset (not included in our seed dictionary) as query terms and retrieve their five nearest neighbors in the English embedding space, thus inducing a bilingual lexicon from the cross-lingual embedding space \cite{bli}. We manually evaluate this dictionary for correct literal translations and report performance in terms of $\mathrm{Precision}@5$: the ratio of query words for which the correct translation is among the word's five nearest neighbors in the target space \cite{lample2018word}. The equation is defined as:
\begin{align*}
\mathrm{Precision}@k = \frac{{{N@k}}}{{{N}}}
\end{align*}

\noindent where $N@k$ is the number of pairs with the correct literal translation in top $k$ nearest neighbors and $N$ is the total number of pairs.

The result is 68\% (i.e. 68 of 100 query words were correctly mapped), which indicates that (a) the global geometry of the two embedding spaces is indeed similar and VecMap has successfully aligned them using a seed lexicon of just 15 word pairs; and that (b) in the majority of the cases there is a 1:1 match between the Chinese and English words. More interesting, however, are the  32  words without a literal match. Here we find that 26 map onto what can be considered a cultural equivalent, while the other six can be considered accidental errors (due to lacking quality in the monolingual embeddings and/or inaccuracies in the alignment). We provide qualitative examples in Table~\ref{tab:bilingual_lexicon_examples}.

\begin{table}[t]
\resizebox{0.49\textwidth}{!}{
        \begin{CJK*}{UTF8}{gbsn}
		\begin{tabular}{lll}
			\toprule
			Source  & Target & Nearest Neighbors   \\
			\midrule
                
                
                \bkcolorliteral{\xpinyin*{水果}}     & fruit           & \underline{fruit}, fruits, kiwi, strawberry, seasonal       \\
                \bkcolorwrong{\xpinyin*{沙拉}}	& salad & feta, lebanese, bruschetta, tabbouleh, caesar   \\
                \bkcolordiff{\xpinyin*{豆腐}}     & tofu            & boiled, \underline{ham}, sausage, bacon, kielbasi           \\
                \bkcolordiff{\xpinyin*{淀粉}}	& starch	& \underline{flour}, beaten, salt, shortening, pwdr \\
                \bkcolordiff{\xpinyin*{筷子} }    & chopstick       & \underline{fork}, spatula, toothpick, wooden, knives \\
                \bkcolordiff{\xpinyin*{蒸}}	& steam	& \underline{bake}, 350, pans, boil, oblong \\
                 
    \bottomrule
	\end{tabular}
    \end{CJK*}}
	\caption{\label{tab:bilingual_lexicon_examples} Top-5 examples from bilingual lexicon induction with \underline{underlined} \bkcolorliteral{literal matches}, \bkcolorwrong{mismatches}, and matches that can be attributed to \bkcolordiff{cultural differences}.}
\end{table}

A successful word match can be exemplified by \begin{CJK}{UTF8}{gbsn}\xpinyin*{水果}\end{CJK} `fruit', which correctly aligns with its English equivalent `fruit' among the top five nearest neighbors. An instance of an inadvertent misalignment, however, can be observed with \begin{CJK}{UTF8}{gbsn}\xpinyin*{沙拉}\end{CJK} `salad'. It is mapped closer to salad ingredients, other side dishes, and particular salad types, rather than precisely corresponding to the English term `salad'.

Certain instances of misalignment can be attributed to cultural differences between English and Chinese culinary practices. Take for instance the ingredient \begin{CJK}{UTF8}{gbsn}\xpinyin*{豆腐}\end{CJK} `tofu', a staple protein source in Chinese cuisine, which aligns with `ham', `sausage', and `bacon'---protein-rich food items prevalent in English-speaking cuisines. Similarly, \begin{CJK}{UTF8}{gbsn}\xpinyin*{淀粉}\end{CJK} `starch' is matched with `flour'. In terms of kitchen utensils, \begin{CJK}{UTF8}{gbsn}\xpinyin*{筷子}\end{CJK} `chopsticks' corresponds to `fork', `spatula', and `toothpick', which perform comparable functions in Western culinary settings. Furthermore, the cooking technique \begin{CJK}{UTF8}{gbsn}\xpinyin*{蒸}\end{CJK} `steam' maps onto `bake', a heat-processing method more frequently used in English recipes. These examples underscore the cultural discrepancies between English and Chinese recipes, emphasizing that recipe adaptation goes beyond mere translation.

\section{Cross-cultural Recipe Adaptation Task}\label{sec:approach}

We propose the task of cross-cultural recipe adaptation, which extends the task of machine translation with the requirement of divergence from the source text semantics in order to address cultural differences in the target culture. While translation studies have long considered culture \cite{bassnett2007culture}, this is not yet explored in machine translation.
Our matched cross-lingual corpora allow us to inform recipe adaptation by both language and culture simultaneously.
In \S\ref{sec:experiments} we adopt an end-to-end sequence-to-sequence approach to the task to establish a set of baselines since this is the dominant approach in machine translation.

The evaluation of cultural adaptation should prioritize meaning preservation while allowing divergences in meaning as long as they stem from cross-cultural differences. This subjective criterion is challenging to implement, as cross-cultural differences, and by extension, the task itself, are not well-defined. As common in text generation tasks, we first adopt reference-based automatic evaluation metrics (\S\ref{sec:autoeval}). Furthermore, to capture structural similarity between references and predictions, we employ meaning representations for evaluation (\S\ref{sec:structeval}). Crucially, since 
reference-based metrics are often unreliable for subjective tasks \citep{reiter-2018-structured}, we additionally perform human evaluation (\S\ref{sec:humaneval}).

\subsection{Surface-based Automatic Evaluation}\label{sec:autoeval}
We use various metrics to assess the similarity between the generated and reference recipes. We use three overlap-based metrics: BLEU \cite{papineni-etal-2002-bleu}, a precision-oriented metric based on token $n$-gram overlap and commonly used in machine translation evaluation, ChrF \cite{popovic-2015-chrf}, a character-level F-score metric that does not depend on tokenization,\footnote{\cy{For BLEU and ChrF, we use \href{https://github.com/mjpost/sacrebleu}{\texttt{SacreBLEU}} \cite{post-2018-call} version 2.3.1 with default parameter settings.}} and ROUGE-L \cite{lin-2004-rouge}, a recall-oriented metric based on longest common subsequences and widely used in summarization evaluation;\footnote{For evaluation, we replace newlines with spaces in all reference and generated recipes. We segment Chinese text to words with \href{https://github.com/fxsjy/jieba}{\texttt{jieba}}.} and one representation-based metric, BERTScore \cite{zhang2019bertscore}, based on cosine similarity of contextualized token embeddings\footnote{We rely on \href{https://huggingface.co/bert-base-uncased}{\texttt{bert-base-uncased}} for representing English text and \href{https://huggingface.co/bert-base-chinese}{\texttt{bert-base-chinese}} for Chinese text.} and shown to correlate better with human judgments than the above metrics in various tasks.

\subsection{Structure-aware Automatic Evaluation}\label{sec:structeval}
Standard metrics may not effectively capture semantic similarity between texts due to sensitivity to surface form. To address this, we employ graph representations, a favored choice for capturing the flow of cooking actions, tool usage, and ingredient transformations in recipes \cite{mori2014flow,kiddon2015mise,jermsurawong2015predicting,yamakata2016method}. These allow for an examination of structural differences influenced by language and culture \cite{wein-etal-2022-effect}.
Here, we leverage Abstract Meaning Representation \cite[AMR;][]{banarescu2013abstract}, a general-purpose graph meaning representation, to represent recipes.

To generate AMR graphs, we employ \texttt{XAMR} \cite{cai-etal-2021-multilingual-amr},\footnote{We use the trained AMR parser model from \url{https://github.com/jcyk/XAMR}.} a state-of-the-art cross-lingual AMR parser that can parse text from five different languages into their corresponding AMR graphs. It is based on a sequence-to-sequence model, utilizing mBART \cite{10.1162/tacl_a_00343} for both encoder and decoder initialization. 

To assess the similarity between model-generated and reference texts' AMRs, we use the \textit{Smatch} metric \cite{cai2013smatch}, which aligns both graphs and computes the F1 score that measures normalized triple overlap.

\subsection{Human Evaluation}\label{sec:humaneval}
While the above automatic metrics provide quantifiable results, they inherently suffer from the limitation of depending on a fixed reference set. In reality, there exist multiple legitimate ways to adapt a recipe. To address this, we propose four criteria for human evaluation, which we conduct on the gold-standard test set. 

We have evaluators assess the outputs from all methods, including the human-written adaptations, on four dimensions key to the cultural adaptation of recipes: (1) \textit{Grammar}---The generated recipe is grammatically sound and fluent; (2) \textit{Consistency}---The output aligns with the format of a fully executable recipe encompassing coherent title, ingredients, and cooking steps; (3) \textit{Preservation}---The adapted recipe largely retains the essence of the source recipe, producing a dish akin to the original; (4) \textit{Cultural Appropriateness}---The generated recipe integrates well with the target cooking culture, aligning with the evaluator's culinary knowledge and recipe style expectations. Evaluators mark each dimension on a 7-point Likert scale \cite{likert1932technique}, where a higher score indicates superior performance. A single evaluator rates each recipe pair separately and independently.

\paragraph{Crowdsourcing Evaluation.}  We recruit evaluators on Prolific\footnote{\url{https://www.prolific.co/}} and deploy our evaluation platform on the same in-house web application used for human recipe writing (\S\ref{sec:humanadap}).
To ensure the evaluation validity, we require participants to be native speakers of the target language and proficient in the source language for each adaptation direction. Additionally, participants must successfully undergo a comprehension check, guided by our evaluation tutorial. Each evaluator is required to evaluate two example recipes for the comprehension check and three recipes for our tasks. This rigorous screening process secures the reliability and accuracy of the evaluations conducted for our study.


\section{Experiments}\label{sec:experiments}

Here we describe our recipe adaptation experiments and results, using the CulturalRecipes dataset introduced in \S\ref{sec:dataset}.
Due to their success in machine translation, we experiment with three end-to-end sequence-to-sequence classes of models to adapt recipes across cultures: (finetuned) machine translation models, finetuned multilingual encoder-decoder models, and prompt-based (zero-shot) multilingual language modeling. Additionally, we evaluate the automatic matching approach used in our dataset construction. These will serve as baselines for future work on this task.

\subsection{Experimental Setup}
We use our silver training set for finetuning in each direction and evaluate on both the silver and gold test sets.
We represent a recipe as a concatenation of title, ingredients, and steps, each section prefixed with a heading (`\texttt{Title:}', `\texttt{Ingredients:}' and `\texttt{Steps:}', for both English and Chinese recipes).\footnote{We treat these headings as language-invariant meta-text, which is removed in post-processing prior to evaluation.}

\paragraph{Automatic matching.}
Since the source recipes used in the creation of the gold-standard test set are a subsample of the ones found in the silver-standard test set, we have matches for them in the target language retrieved based on title similarity (see \S\ref{sec:silver} for a reminder of how the silver-standard test set was constructed). We evaluate these retrieved matches against the gold-standard human-written references, to determine whether title-based retrieval is a viable method for recipe adaptation.

\paragraph{Machine translation.}
Recognizing the intrinsic translation component of recipe adaptation between languages, we leverage pre-trained machine translation systems in our experiments.
We experiment with \texttt{opus-mt} models \citep{TiedemannThottingal:EAMT2020},\footnote{\texttt{Helsinki-NLP/opus-mt-\{\href{https://huggingface.co/Helsinki-NLP/opus-mt-zh-en}{zh-en}/\href{https://huggingface.co/Helsinki-NLP/opus-mt-en-zh}{en-zh}\}}} which show a strong performance in machine translation.
We first evaluate them in zero-shot mode (MT-zs), that is, purely as machine translation models, and additionally after finetuning using our training and validation sets (MT-ft). 

\paragraph{Multilingual language modeling.}
We finetune multilingual encoder-decoder pre-trained language models on the CulturalRecipes dataset. Such models perform well on translation tasks  \cite{https://doi.org/10.48550/arxiv.2008.00401} and are generally trained on abundant monolingual as well as parallel data, so they could prove more suitable for the recipe domain and for our ultimate goal, recipe adaptation.
We choose mT5-base \cite{xue-etal-2021-mt5},\footnote{\href{https://huggingface.co/google/mt5-base}{\texttt{google/mt5-base}}} a multilingual multitask text-to-text transformer pre-trained on a Common Crawl-based dataset containing 101 languages, and mBART50 \cite{https://doi.org/10.48550/arxiv.2008.00401},\footnote{\href{https://huggingface.co/facebook/mbart-large-50}{\texttt{facebook/mbart-large-50}}} a variant of mBART \cite{liu-etal-2020-multilingual-denoising} based on a multilingual autoencoder finetuned for machine translation.

\paragraph{Prompting LLMs.}
Building on the remarkable performance of Multilingual LLMs in zero-shot translation without additional finetuning or in-context learning \cite{https://doi.org/10.48550/arxiv.2106.13627}, we explore their recipe translation and adaptation capabilities.

We use BLOOM \citep{scao2022bloom}, an LLM trained on the multilingual ROOTS corpus \cite{laurenccon2022bigscience}.\footnote{\href{https://huggingface.co/bigscience/bloom-7b1}{\texttt{bigscience/bloom-7b1}}, a 7B-parameter model with a 2k-token length limit. Preliminary experiments showed poor results with \href{https://huggingface.co/bigscience/bloomz-7b1}{BLOOMZ-7B}, \href{https://huggingface.co/bigscience/mt0-xxl-mt}{mT0-xxl-mt} and \href{https://huggingface.co/google/flan-t5-xxl}{FLAN-T5-xxl} \citep{chung2022scaling}, which are finetuned on multitask multilingual prompts \citep{muennighoff2022crosslingual}---they are biased towards short outputs, prevalent in their training tasks.} Using the ROOTS search tool \cite{piktus-etal-2023-roots}, we find it does not contain our recipe corpora. As BLOOM is an autoregressive language model trained to continue text, we prompt as follows for English$\to$Chinese:
\begin{myquote}{0.1in}
\small\texttt{[English recipe] \begin{CJK}{UTF8}{gbsn}\xpinyin*{中文菜谱，适合中国人的:}\end{CJK}}
\end{myquote}
and for Chinese$\to$English:
\begin{myquote}{0.1in}
\small\texttt{[Chinese recipe] Recipe in English, adapted to an English-speaking audience:}
\end{myquote}


\cy{Further, we experiment with GPT-4 \cite{openai2023gpt4},\footnote{\texttt{gpt-4-0314} via the OpenAI API (8k-token length limit).} and ChatGLM2 \cite{zeng2022glm, du2022glm},\footnote{Accessed via \href{https://github.com/lm-sys/FastChat}{FastChat (ChatGLM2-6B)}.} state-of-the-art multilingual and Chinese instruction-tuned LLMs \cite{ouyang2022training}. 
While they have likely been trained on both our recipe corpora (\S\ref{sec:sources}), they do not benefit from our matching procedure (\S\ref{sec:silver}) or our newly-written human-adapted recipes (\S\ref{sec:humanadap}).}
We prompt them as follows for English$\to$Chinese:
\begin{myquote}{0.1in}
\small\texttt{Convert the provided English recipe into a Chinese recipe so that it fits within Chinese cooking culture, is consistent with Chinese cooking knowledge, and meets a Chinese recipe's style. [English recipe]}
\end{myquote}
and for Chinese$\to$English:
\begin{myquote}{0.1in}
\small\texttt{Convert the provided Chinese recipe into an English recipe so that it fits within Western cooking culture, is consistent with Western cooking knowledge, and meets a Western recipe's style. [Chinese recipe]}
\end{myquote}

\paragraph{Technical details.} 
For finetuning, we use a batch size of 64 for MT-ft and 32 for mT5-base and mBART50; and a learning rate of 1e-4.\footnote{Selected among the learning rates \{1e-5, 1e-4\} for MT-ft, \{5e-5, 1e-4\} for mT5-base and mBART50; and batch sizes \{64, 128\} for MT-ft and \{32, 64\} for mT5-base and mBART50.} We set the maximum sequence length to 512 tokens and finetune models for 30 epochs with early stopping after 5 epochs of no improvement in BLEU on the silver validation set. We use two 40GB A100 GPUs for finetuning mT5 and mBART50 and a single one for finetuning MT-ft and for prompting BLOOM. We use the default settings for GPT-4. For ChatGLM2 we set the temperature to 0.7 and the maximum sequence length to 1024 tokens. For generation with all other models, we use a beam of size 3 and a repetition penalty of 1.2; we prevent repeated occurrences of any n-gram of length $\geq$ 5.

\begin{table}[t]
\begin{spacing}{1}
\centering
\resizebox{\linewidth}{!}{
\begin{tabular}{l|ccccc|c}
\toprule 
Method & \textbf{\small BLEU} &\textbf{\small ChrF}  &\textbf{\small R-L} & \textbf{\small B-Sc} & \textbf{\small Smatch} &\textbf{\small \# Tok.} \\
\midrule
\multicolumn{7}{c}{\textbf{Chinese $\rightarrow$ English}} \\
\midrule
MT-zs & 6.8 & 28.7 & 12.0 & 54.0 & 23.7 & 82.4 \\
MT-ft & \textbf{68.9} & \textbf{43.8} & \textbf{22.3} & \textbf{64.6} & \textbf{33.1} & 98.7 \\
mT5  & 60.0 & 37.2 & 19.5 & 62.9 & 31.0 & 85.2 \\
mBART50  & 44.5 & 36.0 & 21.0 & 63.4 & 32.1 & 89.9 \\
\midrule
\multicolumn{7}{c}{\textbf{English $\rightarrow$ Chinese}} \\
\midrule
MT-zs & 2.6 & 9.3 & 49.7 & 62.4 & 20.6 & 110.6 \\
MT-ft & 38.5 & \textbf{37.1} & 54.5 & 71.4 & 26.8 & 91.4 \\
mT5 & \textbf{39.2} & 36.3 & 54.9 & \textbf{71.9} & \textbf{27.0} & 82.1 \\
mBART50 & 30.5 & 32.9 & \textbf{56.2} & 71.1 & 25.5 & 103.2 \\
\bottomrule
\end{tabular}}
\end{spacing}
\caption{\label{tab:auto_results_silver} Automated evaluation results on the silver test sets using reference-based metrics: SacreBLEU (BLEU), ChrF, R-L (ROUGE), B-Sc (BERTScore)---all token-based, and \textit{Smatch}---a structure-aware metric assessing AMR graph similarity. Higher scores indicate better performance on all metrics.}
\end{table}

\begin{table}[t]
\begin{spacing}{1}
\resizebox{\linewidth}{!}{
\begin{tabular}{l|ccccc|c}
\toprule 
Method & \textbf{\small BLEU} &\textbf{\small ChrF}  &\textbf{\small R-L} & \textbf{\small B-Sc} & \textbf{\small Smatch} &\textbf{\small \# Tok.} \\
\midrule
\multicolumn{7}{c}{\textbf{Chinese $\rightarrow$ English}} \\
\midrule
MT-zs$^{\dagger}$ &   5.3   & 29.1 & 22.4  &   59.4  & 30.6 &   77.5   \\
MT-ft &   \textbf{28.0}   & 42.5  &  19.6 &   59.9  & 28.1 &      103.6     \\
mT5  &   14.0   & 31.6  & 17.8  &  59.5  & 25.5 &    87.4     \\
mBART50  &  10.2   & 33.9  & 19.7 &  60.5 & 27.3 & 93.2   \\
BLOOM$^{\dagger}$ &   22.3   & 48.3  &   29.5 &  62.5   & \textbf{33.7} &  110.0    \\
ChatGLM2 & 18.3 & 41.8 & 26.8 & 61.9 & 28.8 & 174.3 \\
GPT-4$^{\dagger}$ & \textbf{28.0}   & \textbf{50.3}  &  \textbf{30.8} &   \textbf{66.5}  & 33.4 &   216.6   \\
Retrieval$^{\dagger}$ & 16.8 & 37.8 & 20.5 & 61.7 & 26.6 & 150.7 \\
\midrule
\multicolumn{7}{c}{\textbf{English $\rightarrow$ Chinese}} \\
\midrule
MT-zs$^{\dagger}$ &  10.6   & 6.9  &  60.8  &   69.8   & 29.4 &     108.0   \\
MT-ft &  13.6   & 28.3 &  53.8  &   70.5   & 24.5 &     88.5    \\ 
mT5 &   16.6   & 28.1 &  53.4  &   70.7   & 25.3 &     78.6      \\
mBART50 & 11.8   & 25.4  &  54.8  &   69.7   & 23.5 &     100.3       \\ 
BLOOM$^{\dagger}$ &    20.0   & 11.5  &  50.8  &   66.4   & 28.6 &     154.7     \\
ChatGLM2 & 22.4 & 11.0 & 54.3 & 75.2 & 28.8 & 153.2 \\
GPT-4$^{\dagger}$ &  21.1   & 21.9  &  \textbf{61.0 }  &   \textbf{77.8}   & \textbf{29.6} &     213.3    \\
Retrieval$^{\dagger}$ & \textbf{32.8} & \textbf{33.6} & 52.9 & 68.4 & 25.0 & 130.3\\
\bottomrule
\end{tabular}}
\end{spacing}
\caption{\label{tab:auto_results_human} Automatic reference-based evaluation results on the gold-standard human test sets. 
$^{\dagger}$ indicates methods without training for the task (zero-shot). 
}
\vspace{-2mm}
\end{table}

\subsection{Results}
\label{sec:results}

\paragraph{Automatic evaluation on the silver test sets.}

As presented in Table~\ref{tab:auto_results_silver}, we restrict our evaluation on the silver-standard test set to finetuned methods,\footnote{We include MT-zs as a reference point to observe the gains from finetuning this model to obtain MT-ft.} as a sanity check for their quality under conditions resembling their training setting. We discern that finetuning the MT model considerably improves its performance across all metrics and in both adaptation directions. In Chinese$\rightarrow$English, MT-zs emerges as the optimal foundation for finetuning, outperforming the other two methods, mT5, and mBART50, across all metrics. However, English$\rightarrow$Chinese displays mixed outcomes, with diverse models excelling in different criteria.
Structure-aware automatic evaluation results generally match other automatic results: MT-ft performs best on Chinese$\to$English, while mT5-base performs best on English$\to$Chinese.

\paragraph{Automatic evaluation on the gold test sets.}

Moving to the gold-standard test set results in Table~\ref{tab:auto_results_human}, we gain further intriguing insights. The significant performance gap between MT-zs and MT-ft reemphasizes that the recipe pairs in our dataset are not merely translations of each other. Moreover, it underscores the systematic patterns in the matched pairs within our training corpus (reflecting the cultural adaptation of recipes) can indeed be learned via finetuning on retrieved recipes. In this scenario, the LLMs BLOOM\cy{, ChatGLM2} and GPT-4 outperform the finetuned methods. Particularly in the Chinese$\rightarrow$English direction, LLMs consistently match or surpass the performance of the next best finetuned approach.
Notably, a comparison of the average length of model predictions shows a tendency of LLMs to produce longer predictions than their counterparts, with GPT-4 generating double the number of tokens compared to other methods.
Interestingly, the retrieval method scores are comparable to the finetuned models in both directions and sometimes even surpass them. Despite this, LLMs continue to prove more effective overall.
\textit{Smatch} scores show performance differences consistent with BERTScore across models for both silver and gold-standard test sets, with the exception that BLOOM slightly outperforms GPT-4 in Chinese$\rightarrow$English.

\begin{table}[t]
    \centering
    \small
    \resizebox{\linewidth}{!}{%
    \setlength{\tabcolsep}{9pt}
    \begin{tabular}{l|cccc}
    \toprule
    Method & \textbf{GRA} & \textbf{CON} & \textbf{PRE} & \textbf{CUL} \\
    \midrule
    \multicolumn{5}{c}{\textbf{Chinese $\rightarrow$ English} ($n=25$)} \\
    \midrule
    MT-zs &    2.6  \scriptsize{$\pm$1.5} &   2.4  \scriptsize{$\pm$1.7} &   2.3  \scriptsize{$\pm$1.4} &      2.7     \scriptsize{$\pm$1.6}    \\
    MT-ft &  4.5  \scriptsize{$\pm$1.8} &   3.7  \scriptsize{$\pm$2.0} &   3.0  \scriptsize{$\pm$2.1} &      4.3     \scriptsize{$\pm$2.1}  \\
    mT5 &  4.1  \scriptsize{$\pm$2.1} &   3.8  \scriptsize{$\pm$2.1} &   3.2  \scriptsize{$\pm$2.2} &      3.7     \scriptsize{$\pm$2.2}   \\
    BLOOM & 3.3  \scriptsize{$\pm$2.0} &   3.3  \scriptsize{$\pm$2.0} &   3.4  \scriptsize{$\pm$2.0} &      2.8     \scriptsize{$\pm$1.8}  \\
    \cy{ChatGLM2} & 4.1 \scriptsize{$\pm$2.4} &    4.3 \scriptsize{$\pm$2.2 } &    4.6 \scriptsize{$\pm$2.1} &     4.0  \scriptsize{$\pm$2.3 } \\
    GPT-4 & \textbf{6.0}  \scriptsize{$\pm$1.2} &   \textbf{6.1}  \scriptsize{$\pm$1.3} &   \textbf{5.9}  \scriptsize{$\pm$1.0} &      \textbf{6.0}     \scriptsize{$\pm$1.2}   \\
    Human & 4.2  \scriptsize{$\pm$2.1} &   4.4  \scriptsize{$\pm$1.9} &   4.5  \scriptsize{$\pm$1.9} &      4.6     \scriptsize{$\pm$1.9}  \\
    Retrieval & 5.1  \scriptsize{$\pm$1.7} &   4.9  \scriptsize{$\pm$2.0} &   4.3  \scriptsize{$\pm$2.3} &      3.8     \scriptsize{$\pm$2.0}  \\
  \midrule
      \multicolumn{5}{c}{\textbf{English $\rightarrow$ Chinese} ($n=41$)} \\
    \midrule
    MT-zs & 2.3  \scriptsize{$\pm$1.6} &   2.7  \scriptsize{$\pm$2.0} &   3.5  \scriptsize{$\pm$2.2} &      2.3     \scriptsize{$\pm$1.7}  \\
    MT-ft &  4.8  \scriptsize{$\pm$2.2} &   3.1  \scriptsize{$\pm$2.2} &   2.5  \scriptsize{$\pm$1.9} &      3.2     \scriptsize{$\pm$2.0}  \\
    mT5 &  4.3  \scriptsize{$\pm$2.0} &   3.4  \scriptsize{$\pm$2.1} &   2.8  \scriptsize{$\pm$2.0} &      3.5     \scriptsize{$\pm$1.9}  \\
    BLOOM & 3.8  \scriptsize{$\pm$2.1} &   4.2  \scriptsize{$\pm$2.1} &   4.6  \scriptsize{$\pm$1.9} &      3.0     \scriptsize{$\pm$1.6} \\
    \cy{ChatGLM2} & 5.4  \scriptsize{$\pm$1.7} & \textbf{5.3}  \scriptsize{$\pm$1.7} & \textbf{5.7}  \scriptsize{$\pm$1.6} & 4.1  \scriptsize{$\pm$2.3}  \\
    GPT-4 & 5.3  \scriptsize{$\pm$2.0} &   5.1  \scriptsize{$\pm$2.0} &   5.2  \scriptsize{$\pm$1.9} &      \textbf{4.4}     \scriptsize{$\pm$2.0}  \\
    Human & \textbf{5.8}  \scriptsize{$\pm$1.1} &   5.1  \scriptsize{$\pm$1.9} &   5.5  \scriptsize{$\pm$1.6} &      4.3     \scriptsize{$\pm$1.8}  \\
    Retrieval & 4.5  \scriptsize{$\pm$1.9} &   3.9  \scriptsize{$\pm$2.0} &   3.3  \scriptsize{$\pm$2.0} &      3.5     \scriptsize{$\pm$1.7}  \\
    \bottomrule
    \end{tabular}}
    \caption{Human evaluation results on the gold-standard test sets: average and standard deviation across recipes for each method and metric, ranging from 1 to 7. Note that different participants manually adapted (``Human'') and evaluated the recipes.}
    \label{tb:table-human}
\end{table}


\begin{table}[t]
\begin{spacing}{1}
\resizebox{.95\linewidth}{!}{
\begin{tabular}{c|*{5}{c}}
\toprule
    & \textbf{BLEU}    & \textbf{ChrF}   & \textbf{R-L}   & \textbf{B-Sc} & \textbf{Smatch} \\ \midrule
    \multicolumn{6}{c}{\textbf{Chinese $\rightarrow$ English}} \\
\midrule
\textbf{GRA}  & 0.135 & 0.250* & 0.135 & 0.257* & 0.021 \\
\textbf{COR}  & 0.151 & 0.268* & 0.180 & 0.294* & 0.065 \\
\textbf{PRE}  & 0.174 & 0.312* & 0.261* & 0.260* & 0.176 \\
\textbf{CUL}  & 0.120 & 0.216* & 0.189 & 0.237* & 0.071 \\
\textbf{\textit{avg.}}  & 0.153 & 0.255* & 0.202* & 0.277* & 0.079 \\
\midrule
    \multicolumn{6}{c}{\textbf{English $\rightarrow$ Chinese}} \\
\midrule
\textbf{GRA}  & 0.286* & 0.353* & 0.201* & 0.278* & 0.070 \\
\textbf{COR}  & 0.227* & 0.232* & 0.183* & 0.217* & 0.116 \\
\textbf{PRE}  & 0.268* & 0.180* & 0.218* & 0.247* & 0.124 \\
\textbf{CUL}  & 0.216* & 0.268* & 0.155 & 0.219* & 0.081 \\
\textbf{\textit{avg.}}  & 0.290* & 0.295* & 0.221* & 0.272* & 0.117 \\
\bottomrule
\end{tabular}}
\end{spacing}
\caption{Kendall correlation of human evaluation results with automatic metrics. Statistically significant correlations are marked with *,  with a confidence level of $\alpha$ = 0.05 before adjusting for multiple comparisons using the Bonferroni correction \cite{bonferroni1936teoria}.}
\label{tb:table-human-corr-kendall}
\end{table}

\paragraph{Human evaluation.}
Table~\ref{tb:table-human} showcases the results of human evaluation, with abbreviations GRA, CON, PRE, and CUL representing Grammar, Consistency, Preservation, and Cultural Appropriateness, respectively.\footnote{We exclude mBART50 due to its architectural and performance similarity to mT5.} GPT-4 excels significantly across all metrics in the Chinese$\rightarrow$English direction, even surpassing explicit human adaptation. Recipes retrieved from popular websites are a close second in GRA and CON, reflecting their high quality. However, the targeted adaptations written by humans who were explicitly instructed to adapt the source recipe to the target culture, perform better in PRE and CUL.
For English$\rightarrow$Chinese, \cy{GPT-4 remains the top performer only in CUL, while mT5 parallels the retrieved recipes in this metric. Notably, ChatGLM2 surpasses even human writers in CON and PRE, but not in GRA.}

\paragraph{Correlation of automatic metrics with humans.}
To determine the reliability of automatic metrics in assessing the quality of recipe adaptations, we examine their correlation with human evaluations across the four metrics and their average. We use Kendall correlation, which is the official meta-evaluation metric used by WMT22 metric shared task \cite{freitag-etal-2022-results}.

As illustrated in Table~\ref{tb:table-human-corr-kendall}, all cases exhibit a positive correlation, albeit with varying strengths from weak to moderate, and with inconsistent performance between the two adaptation directions. For Chinese$\rightarrow$English, ChrF and BERTScore indicate the strongest correlation with the average of all criteria. BERTScore further stands out by demonstrating the highest correlation with each individual criterion. On the other hand, for English$\rightarrow$Chinese, BLEU performs comparably well, thus highlighting that the effectiveness of these metrics can vary based on the direction of adaptation. ROUGE-L, however, displays a significantly lower correlation, suggesting its limitations in evaluating recipe adaptations. 
Finally, we observe that \textit{Smatch} is not significantly correlated with human judgments, possibly due to noise introduced by parsing errors.\footnote{Inspecting \texttt{XAMR} outputs, we notice recurrent errors in both languages, likely attributable to the unique recipe genre. Common culinary actions are often incorrectly represented or overlooked: in English, actions like `oil' or `grease' are treated as objects. Similarly in Chinese, many actions are often omitted or associated with unrelated concepts.}

CUL presents the weakest correlation with most automatic metrics, underscoring the current limitations of automated evaluations in assessing the cultural alignment of recipes, and highlighting the essential role of human evaluators. Notably, correlations for English$\rightarrow$Chinese generally exhibit greater strength than Chinese$\rightarrow$English. This discrepancy is likely due to the variation in sample sizes between the two directions.


\section{Analysis and Discussion}\label{sec:discussion}

Our findings reinforce previous research asserting the cultural bias of LLMs---specifically GPT-4---towards Western, English-speaking, U.S. culture, as exemplified in the food domain \cite{cao-etal-2023-assessing,naous2023having,keleg-magdy-2023-dlama,palta-rudinger-2023-fork}. However, our results also offer a more nuanced perspective. While GPT-4 demonstrates an exceptional ability to adapt to Chinese cuisine, its linguistic and semantic capabilities \cy{are outperformed by ChatGLM2} in English$\rightarrow$Chinese. To delve deeper into these intriguing results, this section examines the strategies these models employ in the adaptation task.

\begin{figure}[t]
    \centering
    \includegraphics[width=0.9\linewidth]{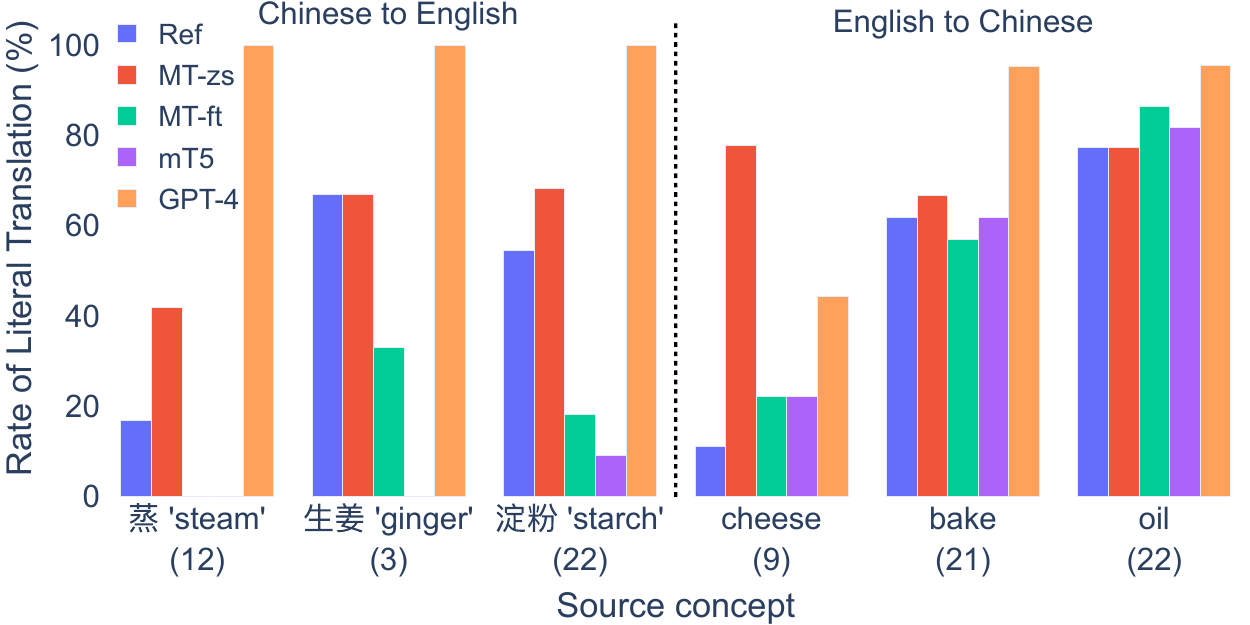}
    \caption{Analysis of the translation of specific concepts by the different models on the silver-standard test data. Ref = retrieved reference. In brackets, we show the number of occurrences of each concept.}
    \label{fig:hyp_test}
\end{figure}

\paragraph{Quantitative analysis.}
Referring back to the analysis from \S\ref{sec:alignment}, we choose a subset of six words and examine how they are handled by four models (MT-zs, MT-ft, and mT5 and GPT-4). Specifically, we measure the rate of literal translation of these concepts by each model, in the context of the recipes from the silver-standard test set of CulturalRecipes.\footnote{We use the silver-standard test set rather than the gold-standard test set for its comparatively larger size.} For instance, in adapting from English to Chinese, we identify \textit{baking} as an English-specific concept. We count the appearances of related terms such as `bake', `roast', `broil', and `oven' in English source recipes, denoted as $c_{source}$. For each instance, we tally the occurrences of the direct translation, \begin{CJK}{UTF8}{gbsn}\xpinyin*{烤}\end{CJK}, in the corresponding Chinese recipes, denoted as $c_{target}$, from either model predictions or retrieved references. We calculate the literal translation rate as $\frac{c_{target}}{c_{source}}$. Figure~\ref{fig:hyp_test} visualizes the results for five culturally-specific concepts and a universally applicable concept, `oil'. 

We include `oil' as a sanity check and indeed see that the literal rate of translation is high in both the references and in all model predictions.

The references show a low to medium rate of literal translations for the remaining five concepts, confirming their cultural specificity. MT-zs often translates these concepts literally, as could be expected from a machine translation model designed for near-literal translation---the difference is especially noticeable for the concepts `steam' and `cheese'. The finetuned models MT-ft and mT5, on the other hand, learn to avoid literal translation, presumably opting for culturally-appropriate alternatives instead---for `steam' for example none of the 12 occurrences of the concept in the source Chinese recipe are literally translated in the predictions of MT-ft and mT5.

An interesting trend emerges in GPT-4 predictions, where literal translations are found at a high rate for all concepts, often close to 100\%. While this seems counter-intuitive considering the goal of adapting the culturally-specific ingredients and cooking methods, in the next section we find that GPT-4 employs a slightly different strategy than just substituting these ingredients and methods. 

\paragraph{Qualitative Analysis.}
We present a qualitative analysis highlighting the adaptation strategies adopted by models, specifically MT-zs, MT-ft, and GPT-4. The analysis centers on the Chinese recipe shown in Figure~\ref{fig:figure1}, with model predictions shown in Table~\ref{tb:case_study}.
The translation from \textbf{MT-zs} directly incorporates Chinese ingredients not common in English recipes, accompanied by numerous spelling and grammatical errors. The prevalence of errors can be attributed to a dearth of recipe domain representations in the machine translation training data of MT-zs.
In contrast, \textbf{MT-ft} offers a notably improved recipe rendition, albeit a wholly different red bean soup from the source recipe. Although this results in minimal content retention, it can be viewed as an extreme cultural adaptation, given the infrequent appearance of sweet red bean soup in Western cuisine. However, MT-ft sporadically manifests consistency errors, exemplified in this case by duplicating beans in the ingredient list and parsley in the steps. These findings confirm that the generation of coherent recipes continues to be a challenging endeavor for sequence-to-sequence models, corroborating the findings of prior work \cite{li-etal-2022-share}.\footnote{Similar behavior is observed in the other sequence-to-sequence models trained on our training set and in the automatically matched (retrieved) recipe.}
\textbf{GPT-4}, on the other hand, generates a recipe more closely aligned with the source than the human-generated reference (refer to Figure~\ref{fig:figure1}). This model also incorporates thoughtful cultural adaptations: it quantifies ingredient amounts, unlike the source which vaguely indicates ``\begin{CJK}{UTF8}{gbsn}\xpinyin*{适量}\end{CJK}'' (\textit{moderate amount}), and it provides alternative names or substitutions for uniquely Chinese ingredients. The recipe instructions retain the crucial details from the source recipe, whilst maintaining fluency and appropriateness for Western-style recipes.

\begin{table}[t]
    \centering
    \tiny
    \begin{CJK*}{UTF8}{gbsn}
    \begin{tabularx}{\columnwidth}{@{}X@{}}
    \toprule 
    \textbf{MT-zs}: 
        Title: \bkcolorliteral{Red beans soup}, \reduwave{Ingridients}: \bkcolorliteral{red beans, rice wine}, \reduwave{\bkcolorliteral{leather}} ginger, Steps: \reduwave{Chrysotile, and red beans in the wine}, soaked for eight hours, soaked in the pot, boiled \reduwave{in fire, stired in fire, turned into small fires} for 30 minutes. \\
    \midrule
    \textbf{MT-ft}: 
    Title: Red Bean Soup, Ingredients: \bkcolordiff{2 tablespoons olive oil 1 large onion, chopped 2 garlic cloves, minced 1 teaspoon dried oregano 1 teaspoons dried basil 1 teaspooon dried thyme 1/2 teaspoon salt 1/4 teaspoon pepper 2 cups red kidney beans, rinsed and drained 1 (14 1/2 ounce) can diced tomatoes, undrained \reduwave{1 (15 ounce) can red kidney beans, un-drained} 1 (14 ounce) can beef broth 1 cup} water \bkcolordiff{1/4 cup chopped fresh parsley}, Steps: \bkcolordiff{Heat oil in a large saucepan over medium heat. Add onion and garlic; cook and stir until tender. Stir in oregano, basil, thyme, salt and pepper. Add beans, tomatoes, broth, water and parsley}; bring to a boil. Reduce heat; cover and simmer for 30 minutes or until beans are tender. Remove from heat; let stand for 5 minutes. \bkcolordiff{\reduwave{Stir in parsley.}} \\
    \midrule
    \textbf{GPT-4}:
    Title: \bkcolorliteral{Red Bean Soup} \\Ingredients: \\- \bkcolordiff{1 cup} red beans \bkcolordiff{(adzuki beans)} \\- \bkcolordiff{1/4 cup} rice wine \bkcolordiff{(or substitute with dry sherry)} \\- \bkcolordiff{1-inch piece} of fresh ginger, \bkcolorliteral{with skin} \\Instructions: \\1. Rinse the red beans and place them in a bowl. Thinly slice the ginger and add it to the bowl with the beans. \\2. Pour the rice wine over the beans and ginger, ensuring they are fully submerged. Stir to combine, then cover and let soak for 8 hours or overnight. \\3. After soaking, transfer the beans, ginger, and soaking liquid to a large pot. Bring the mixture to a boil over high heat, stirring occasionally. \\4. Once boiling, reduce the heat to low and let the soup simmer for 30 minutes, stirring occasionally to prevent sticking. \\5. Taste the soup and adjust the seasoning if necessary. Serve hot and enjoy your Red Bean Soup! \\
    \bottomrule
    \end{tabularx}
    \end{CJK*}
    \caption{Case study: English adaptations of the Chinese recipe from Figure~\ref{fig:figure1}, with manually highlighted \reduwave{(spelling, grammar or semantic) errors}, \bkcolordiff{adaptations to cultural differences}, and \bkcolorliteral{failures} to account for such.}
    \label{tb:case_study}
\end{table}

\section{Related Work}\label{sec:related}

\paragraph{Cultural adaptation of text.}
Cultural adaptation overlaps with style transfer, where the goal is to change the style of text while preserving the meaning \cite{jin-etal-2022-deep}.
In addition to style, cultural adaptation also concerns common ground, values and topics of interest \cite{hershcovich-etal-2022-challenges}. Particularly in culture-loaded tasks, it becomes crucial to consider cultural differences~\cite{zhou-etal-2023-cross, zhou2023cultural}.
While semantic divergences are usually treated as errors in machine translation \cite{briakou-carpuat-2021-beyond}, cross-cultural translation often requires adaptations that change the meaning, e.g., by adapting entities \cite{peskov-etal-2021-adapting-entities} or by adding explanations \cite{kementchedjhieva-etal-2020-apposcorpus}.
We share the motivation of this line of work, but for the first time focus on recipes, where cultural adaptation is grounded in clear goals (accessibility to the cook and quality of the resulting dish).

\paragraph{Recipe generation.}
\Citet{10.3389/frai.2020.621577} outline potential cross-disciplinary approaches involving NLP and food science, claiming that the analysis of digital recipes is a promising but challenging task.
\citet{marin2019learning} introduce the Recipe1M dataset (see \S\ref{sec:dataset}) and \citet{h2020recipegpt} finetune GPT-2 \cite{radford2019language} on it to create a large English language model, RecipeGPT, capable of generating cooking instructions from titles and ingredients or ingredients from instructions and titles.
\citet{majumder-etal-2019-generating} introduce a dataset of 180K English recipes from the website Food.com and a neural model to generate recipes according to user preferences inferred from historical interactions.
Contrary to these, we focus on recipe adaptation, where generation is conditioned on a source recipe.

\paragraph{Recipe adaptation.}
\citet{donatelli-etal-2021-aligning} align recipes for the same dish on the action level using recipe graphs \cite{yamakata2016method}, aiming to adapt recipes to users of different levels of expertise.
\citet{morales-garzon-etal-2021-semantic,9351987,10.1007/978-3-031-08974-9_24} propose an unsupervised method to adapt recipes according to dietary preferences by proposing ingredient substitutions using domain-specific word and sentence embeddings. However, they do not modify the recipe steps beyond simple ingredient substitution.
\citet{li-etal-2022-share} build a dataset of 83K automatically-matched recipe pairs for the task of editing recipes to satisfy dietary restrictions. They train a supervised model to perform controlled generation, outperforming RecipeGPT. They identify the remaining challenge of ``controllable recipe editing using more subtle traits such as cuisines (e.g., making a Chinese version of meatloaf)'', which we address here.
\citet{antognini-etal-2023-assistive}, in contrast, propose addressing the same task \textit{without} paired data, utilizing an unsupervised critiquing module and also outperforming RecipeGPT in both automatic and human evaluation.
\citet{liu-etal-2022-counterfactual} present a dataset of 1.5M Chinese recipes and evaluate compositional generalization in neural models in the task of counterfactual generation of recipes with substituted ingredients.
They find recipe adaptation to be a challenging task: language models often generate incoherent recipes or fail to satisfy the stated constraints. In contrast, we find that after finetuning pre-trained models on our dataset, the models succeed in the task of cultural adaptation.

\section{Conclusion and Future Work}\label{sec:conclusion}

In this work, we studied the task of adapting cooking recipes across cultures. We identified dimensions relevant to this task through a data-driven analysis, including differences in ingredients, tools, methods, and measurement units. We introduced CulturalRecipes, a dataset of paired Chinese and English recipes, and evaluated various adaptation methods. Through our experiments and analysis, we show that models can learn to consider cultural aspects, including style, when adapting recipes across cultures, with some challenges remaining in the level of detail and consistency between the different components of a recipe.

We envision our dataset and baselines will be useful for both downstream applications and further studies of cultural adaptation within and beyond NLP. Automatically adapting recipes from one culture to another could facilitate cross-cultural cross-pollination and broaden the horizons of potential users, serving as a bridge between people through food, and being useful to both novice and experienced cooks. Furthermore, our dataset is a challenging benchmark for language models: besides the complex compositional generalization ability required for recipe adaptation \cite{liu-etal-2022-counterfactual}, it assesses the ability of multilingual language models to adapt to target cultural characteristics, and to construct well-formed and faithful recipes. Lastly, our cross-cultural comparative analysis can be extended to sociological and anthropological research.

\paragraph{Future work.} As acknowledged in \S\ref{sec:differences}, the cultural categories we assume are highly simplistic. Future work will expand our datasets to treat finer-grained differences, as well as broaden it to more languages and cultures. It will further investigate the factors that impact recipe adaptation and develop more sophisticated modeling approaches to consider them, beyond the sequence-to-sequence approaches we experimented with here. Finally, our dataset can provide a starting point for related tasks, including recipe classification and retrieval.

Cultural categorization can be a sensitive topic so we have been careful to approach it with respect for the communities involved; we encourage future research in the area to maintain this practice. We hope that our research can contribute to a greater understanding and appreciation of diverse cultural traditions and practices related to food and cooking.

\section*{Acknowledgments}
The authors extend their sincere gratitude to the reviewers and action editors for their invaluable feedback, which significantly contributed to the improvement of this work. Special thanks are also due to Laura Cabello and Nicolas Garneau for their insightful comments and to Qinghua Zhao and Jingcun Huang for their valuable assistance during our initial human evaluations. The authors gratefully acknowledge the HPC RIVR consortium (www.hpc-rivr.si) and EuroHPC JU (eurohpc-ju.europa.eu) for funding this research by providing computing resources of the HPC system Vega at the Institute of Information Science (www.izum.si). Yong Cao and Li Zhou gratefully acknowledge financial support from China Scholarship Council. (CSC No. 202206070002 and No. 202206160052).


\bibliography{anthology,custom}

\begin{thebibliography}{70}
\expandafter\ifx\csname natexlab\endcsname\relax\def\natexlab#1{#1}\fi

\bibitem[{Ahn et~al.(2011)Ahn, Ahnert, Bagrow, and
  Barab{\'a}si}]{ahn2011flavor}
Yong-Yeol Ahn, Sebastian~E Ahnert, James~P Bagrow, and Albert-L{\'a}szl{\'o}
  Barab{\'a}si. 2011.
\newblock Flavor network and the principles of food pairing.
\newblock \emph{Scientific reports}, 1(1):196.

\bibitem[{Albala(2012)}]{albala2012three}
Ken Albala. 2012.
\newblock \href {https://books.google.dk/books?id=JuzLHwNw9J0C} {\emph{Three
  World Cuisines: {I}talian, {M}exican, {C}hinese}}.
\newblock Rowman Altamira.

\bibitem[{Antognini et~al.(2023)Antognini, Li, Faltings, and
  McAuley}]{antognini-etal-2023-assistive}
Diego Antognini, Shuyang Li, Boi Faltings, and Julian McAuley. 2023.
\newblock \href {https://aclanthology.org/2023.eacl-main.28} {Assistive recipe
  editing through critiquing}.
\newblock In \emph{Proceedings of the 17th Conference of the European Chapter
  of the Association for Computational Linguistics}, pages 375--384, Dubrovnik,
  Croatia. Association for Computational Linguistics.

\bibitem[{Artetxe et~al.(2017)Artetxe, Labaka, and Agirre}]{artetxe2017acl}
Mikel Artetxe, Gorka Labaka, and Eneko Agirre. 2017.
\newblock Learning bilingual word embeddings with (almost) no bilingual data.
\newblock In \emph{Proceedings of the 55th Annual Meeting of the Association
  for Computational Linguistics (Volume 1: Long Papers)}, pages 451--462.

\bibitem[{Banarescu et~al.(2013)Banarescu, Bonial, Cai, Georgescu, Griffitt,
  Hermjakob, Knight, Koehn, Palmer, and Schneider}]{banarescu2013abstract}
Laura Banarescu, Claire Bonial, Shu Cai, Madalina Georgescu, Kira Griffitt, Ulf
  Hermjakob, Kevin Knight, Philipp Koehn, Martha Palmer, and Nathan Schneider.
  2013.
\newblock Abstract meaning representation for sembanking.
\newblock In \emph{Proceedings of the 7th linguistic annotation workshop and
  interoperability with discourse}, pages 178--186.

\bibitem[{Bassnett(2007)}]{bassnett2007culture}
Susan Bassnett. 2007.
\newblock Culture and translation.
\newblock \emph{A companion to translation studies}, pages 13--23.

\bibitem[{Bie{\'n} et~al.(2020)Bie{\'n}, Gilski, Maciejewska, Taisner,
  Wisniewski, and Lawrynowicz}]{bien-etal-2020-recipenlg}
Micha{\l} Bie{\'n}, Micha{\l} Gilski, Martyna Maciejewska, Wojciech Taisner,
  Dawid Wisniewski, and Agnieszka Lawrynowicz. 2020.
\newblock \href {https://aclanthology.org/2020.inlg-1.4} {{R}ecipe{NLG}: A
  cooking recipes dataset for semi-structured text generation}.
\newblock In \emph{Proceedings of the 13th International Conference on Natural
  Language Generation}, pages 22--28, Dublin, Ireland. Association for
  Computational Linguistics.

\bibitem[{Bonferroni(1936)}]{bonferroni1936teoria}
Carlo Bonferroni. 1936.
\newblock Teoria statistica delle classi e calcolo delle probabilita.
\newblock \emph{Pubblicazioni del R Istituto Superiore di Scienze Economiche e
  Commericiali di Firenze}, 8:3--62.

\bibitem[{Briakou and Carpuat(2021)}]{briakou-carpuat-2021-beyond}
Eleftheria Briakou and Marine Carpuat. 2021.
\newblock \href {https://doi.org/10.18653/v1/2021.acl-long.562} {Beyond noise:
  Mitigating the impact of fine-grained semantic divergences on neural machine
  translation}.
\newblock In \emph{Proceedings of the 59th Annual Meeting of the Association
  for Computational Linguistics and the 11th International Joint Conference on
  Natural Language Processing (Volume 1: Long Papers)}, pages 7236--7249,
  Online. Association for Computational Linguistics.

\bibitem[{Cai et~al.(2021)Cai, Li, Ho, Bing, and
  Lam}]{cai-etal-2021-multilingual-amr}
Deng Cai, Xin Li, Jackie Chun-Sing Ho, Lidong Bing, and Wai Lam. 2021.
\newblock \href {https://doi.org/10.18653/v1/2021.findings-emnlp.237}
  {Multilingual {AMR} parsing with noisy knowledge distillation}.
\newblock In \emph{Findings of the Association for Computational Linguistics:
  EMNLP 2021}, pages 2778--2789, Punta Cana, Dominican Republic. Association
  for Computational Linguistics.

\bibitem[{Cai and Knight(2013)}]{cai2013smatch}
Shu Cai and Kevin Knight. 2013.
\newblock Smatch: an evaluation metric for semantic feature structures.
\newblock In \emph{Proceedings of the 51st Annual Meeting of the Association
  for Computational Linguistics (Volume 2: Short Papers)}, pages 748--752.

\bibitem[{Cao et~al.(2023)Cao, Zhou, Lee, Cabello, Chen, and
  Hershcovich}]{cao-etal-2023-assessing}
Yong Cao, Li~Zhou, Seolhwa Lee, Laura Cabello, Min Chen, and Daniel
  Hershcovich. 2023.
\newblock \href {https://aclanthology.org/2023.c3nlp-1.7} {Assessing
  cross-cultural alignment between {C}hat{GPT} and human societies: An
  empirical study}.
\newblock In \emph{Proceedings of the First Workshop on Cross-Cultural
  Considerations in NLP (C3NLP)}, pages 53--67, Dubrovnik, Croatia. Association
  for Computational Linguistics.

\bibitem[{Chung et~al.(2022)Chung, Hou, Longpre, Zoph, Tay, Fedus, Li, Wang,
  Dehghani, Brahma et~al.}]{chung2022scaling}
Hyung~Won Chung, Le~Hou, Shayne Longpre, Barret Zoph, Yi~Tay, William Fedus,
  Eric Li, Xuezhi Wang, Mostafa Dehghani, Siddhartha Brahma, et~al. 2022.
\newblock Scaling instruction-finetuned language models.
\newblock \emph{arXiv preprint arXiv:2210.11416}.

\bibitem[{Donatelli et~al.(2021)Donatelli, Schmidt, Biswas, K{\"o}hn, Zhai, and
  Koller}]{donatelli-etal-2021-aligning}
Lucia Donatelli, Theresa Schmidt, Debanjali Biswas, Arne K{\"o}hn, Fangzhou
  Zhai, and Alexander Koller. 2021.
\newblock \href {https://doi.org/10.18653/v1/2021.emnlp-main.554} {Aligning
  actions across recipe graphs}.
\newblock In \emph{Proceedings of the 2021 Conference on Empirical Methods in
  Natural Language Processing}, pages 6930--6942, Online and Punta Cana,
  Dominican Republic. Association for Computational Linguistics.

\bibitem[{Du et~al.(2022)Du, Qian, Liu, Ding, Qiu, Yang, and Tang}]{du2022glm}
Zhengxiao Du, Yujie Qian, Xiao Liu, Ming Ding, Jiezhong Qiu, Zhilin Yang, and
  Jie Tang. 2022.
\newblock Glm: General language model pretraining with autoregressive blank
  infilling.
\newblock In \emph{Proceedings of the 60th Annual Meeting of the Association
  for Computational Linguistics (Volume 1: Long Papers)}, pages 320--335.

\bibitem[{van Erp et~al.(2021)van Erp, Reynolds, Maynard, Starke,
  Ibáñez~Martín, Andres, Leite, Alvarez~de Toledo, Schmidt~Rivera, Trattner,
  Brewer, Adriano~Martins, Kluczkovski, Frankowska, Bridle, Levy, Rauber,
  Tereza~da Silva, and Bosma}]{10.3389/frai.2020.621577}
Marieke van Erp, Christian Reynolds, Diana Maynard, Alain Starke, Rebeca
  Ibáñez~Martín, Frederic Andres, Maria C.~A. Leite, Damien Alvarez~de
  Toledo, Ximena Schmidt~Rivera, Christoph Trattner, Steven Brewer, Carla
  Adriano~Martins, Alana Kluczkovski, Angelina Frankowska, Sarah Bridle,
  Renata~Bertazzi Levy, Fernanda Rauber, Jacqueline Tereza~da Silva, and Ulbe
  Bosma. 2021.
\newblock \href {https://doi.org/10.3389/frai.2020.621577} {Using natural
  language processing and artificial intelligence to explore the nutrition and
  sustainability of recipes and food}.
\newblock \emph{Frontiers in Artificial Intelligence}, 3.

\bibitem[{Freitag et~al.(2022)Freitag, Rei, Mathur, Lo, Stewart, Avramidis,
  Kocmi, Foster, Lavie, and Martins}]{freitag-etal-2022-results}
Markus Freitag, Ricardo Rei, Nitika Mathur, Chi-kiu Lo, Craig Stewart,
  Eleftherios Avramidis, Tom Kocmi, George Foster, Alon Lavie, and Andr{\'e}
  F.~T. Martins. 2022.
\newblock \href {https://aclanthology.org/2022.wmt-1.2} {Results of {WMT}22
  metrics shared task: Stop using {BLEU} {--} neural metrics are better and
  more robust}.
\newblock In \emph{Proceedings of the Seventh Conference on Machine Translation
  (WMT)}, pages 46--68, Abu Dhabi, United Arab Emirates (Hybrid). Association
  for Computational Linguistics.

\bibitem[{H.~Lee et~al.(2020)H.~Lee, Shu, Achananuparp, Prasetyo, Liu, Lim, and
  Varshney}]{h2020recipegpt}
Helena H.~Lee, Ke~Shu, Palakorn Achananuparp, Philips~Kokoh Prasetyo, Yue Liu,
  Ee-Peng Lim, and Lav~R Varshney. 2020.
\newblock \href {https://arxiv.org/abs/2003.02498} {Recipe{GPT}: Generative
  pre-training based cooking recipe generation and evaluation system}.
\newblock In \emph{Companion Proceedings of the Web Conference 2020}, pages
  181--184.

\bibitem[{Hershcovich et~al.(2022)Hershcovich, Frank, Lent, de~Lhoneux, Abdou,
  Brandl, Bugliarello, Cabello~Piqueras, Chalkidis, Cui, Fierro, Margatina,
  Rust, and S{\o}gaard}]{hershcovich-etal-2022-challenges}
Daniel Hershcovich, Stella Frank, Heather Lent, Miryam de~Lhoneux, Mostafa
  Abdou, Stephanie Brandl, Emanuele Bugliarello, Laura Cabello~Piqueras, Ilias
  Chalkidis, Ruixiang Cui, Constanza Fierro, Katerina Margatina, Phillip Rust,
  and Anders S{\o}gaard. 2022.
\newblock \href {https://doi.org/10.18653/v1/2022.acl-long.482} {Challenges and
  strategies in cross-cultural {NLP}}.
\newblock In \emph{Proceedings of the 60th Annual Meeting of the Association
  for Computational Linguistics (Volume 1: Long Papers)}, pages 6997--7013,
  Dublin, Ireland. Association for Computational Linguistics.

\bibitem[{Jermsurawong and Habash(2015)}]{jermsurawong2015predicting}
Jermsak Jermsurawong and Nizar Habash. 2015.
\newblock \href {https://doi.org/10.18653/v1/D15-1090} {Predicting the
  structure of cooking recipes}.
\newblock In \emph{Proceedings of the 2015 Conference on Empirical Methods in
  Natural Language Processing}, pages 781--786, Lisbon, Portugal. Association
  for Computational Linguistics.

\bibitem[{Jin et~al.(2022)Jin, Jin, Hu, Vechtomova, and
  Mihalcea}]{jin-etal-2022-deep}
Di~Jin, Zhijing Jin, Zhiting Hu, Olga Vechtomova, and Rada Mihalcea. 2022.
\newblock \href {https://doi.org/10.1162/coli_a_00426} {Deep learning for text
  style transfer: A survey}.
\newblock \emph{Computational Linguistics}, 48(1):155--205.

\bibitem[{Keleg and Magdy(2023)}]{keleg-magdy-2023-dlama}
Amr Keleg and Walid Magdy. 2023.
\newblock \href {https://aclanthology.org/2023.findings-acl.389} {{DLAMA}: A
  framework for curating culturally diverse facts for probing the knowledge of
  pretrained language models}.
\newblock In \emph{Findings of the Association for Computational Linguistics:
  ACL 2023}, pages 6245--6266, Toronto, Canada. Association for Computational
  Linguistics.

\bibitem[{Kementchedjhieva et~al.(2020)Kementchedjhieva, Lu, and
  Tetreault}]{kementchedjhieva-etal-2020-apposcorpus}
Yova Kementchedjhieva, Di~Lu, and Joel Tetreault. 2020.
\newblock \href {https://doi.org/10.18653/v1/2020.coling-main.180} {The
  {A}ppos{C}orpus: a new multilingual, multi-domain dataset for factual
  appositive generation}.
\newblock In \emph{Proceedings of the 28th International Conference on
  Computational Linguistics}, pages 1989--2003, Barcelona, Spain (Online).
  International Committee on Computational Linguistics.

\bibitem[{Kiddon et~al.(2015)Kiddon, Ponnuraj, Zettlemoyer, and
  Choi}]{kiddon2015mise}
Chlo{\'e} Kiddon, Ganesa~Thandavam Ponnuraj, Luke Zettlemoyer, and Yejin Choi.
  2015.
\newblock Mise en place: Unsupervised interpretation of instructional recipes.
\newblock In \emph{Proceedings of the 2015 Conference on Empirical Methods in
  Natural Language Processing}, pages 982--992.

\bibitem[{Lample et~al.(2018)Lample, Conneau, Ranzato, Denoyer, and
  Jégou}]{lample2018word}
Guillaume Lample, Alexis Conneau, Marc'Aurelio Ranzato, Ludovic Denoyer, and
  Hervé Jégou. 2018.
\newblock \href {https://openreview.net/forum?id=H196sainb} {Word translation
  without parallel data}.
\newblock In \emph{International Conference on Learning Representations}.

\bibitem[{Lauren{\c{c}}on et~al.(2022)Lauren{\c{c}}on, Saulnier, Wang, Akiki,
  Villanova~del Moral, Le~Scao, Von~Werra, Mou, Gonz{\'a}lez~Ponferrada, Nguyen
  et~al.}]{laurenccon2022bigscience}
Hugo Lauren{\c{c}}on, Lucile Saulnier, Thomas Wang, Christopher Akiki, Albert
  Villanova~del Moral, Teven Le~Scao, Leandro Von~Werra, Chenghao Mou, Eduardo
  Gonz{\'a}lez~Ponferrada, Huu Nguyen, et~al. 2022.
\newblock The {B}ig{S}cience {ROOTS} corpus: A 1.6{TB} composite multilingual
  dataset.
\newblock \emph{Advances in Neural Information Processing Systems},
  35:31809--31826.

\bibitem[{Li et~al.(2022)Li, Li, Ni, and McAuley}]{li-etal-2022-share}
Shuyang Li, Yufei Li, Jianmo Ni, and Julian McAuley. 2022.
\newblock \href {https://aclanthology.org/2022.emnlp-main.761} {{SHARE}: a
  system for hierarchical assistive recipe editing}.
\newblock In \emph{Proceedings of the 2022 Conference on Empirical Methods in
  Natural Language Processing}, pages 11077--11090, Abu Dhabi, United Arab
  Emirates. Association for Computational Linguistics.

\bibitem[{Likert(1932)}]{likert1932technique}
Rensis Likert. 1932.
\newblock A technique for the measurement of attitudes.
\newblock \emph{Archives of psychology}.

\bibitem[{Lin et~al.(2020)Lin, Rao, Celikyilmaz, Nouri, Brockett, Dey, and
  Dolan}]{lin-etal-2020-recipe}
Angela Lin, Sudha Rao, Asli Celikyilmaz, Elnaz Nouri, Chris Brockett,
  Debadeepta Dey, and Bill Dolan. 2020.
\newblock \href {https://doi.org/10.18653/v1/2020.acl-main.440} {A recipe for
  creating multimodal aligned datasets for sequential tasks}.
\newblock In \emph{Proceedings of the 58th Annual Meeting of the Association
  for Computational Linguistics}, pages 4871--4884, Online. Association for
  Computational Linguistics.

\bibitem[{Lin(2004)}]{lin-2004-rouge}
Chin-Yew Lin. 2004.
\newblock \href {https://aclanthology.org/W04-1013} {{ROUGE}: A package for
  automatic evaluation of summaries}.
\newblock In \emph{Text Summarization Branches Out}, pages 74--81, Barcelona,
  Spain. Association for Computational Linguistics.

\bibitem[{Liu et~al.(2022)Liu, Feng, Tang, Hu, and
  Zhao}]{liu-etal-2022-counterfactual}
Xiao Liu, Yansong Feng, Jizhi Tang, Chengang Hu, and Dongyan Zhao. 2022.
\newblock \href {https://aclanthology.org/2022.emnlp-main.497} {Counterfactual
  recipe generation: Exploring compositional generalization in a realistic
  scenario}.
\newblock In \emph{Proceedings of the 2022 Conference on Empirical Methods in
  Natural Language Processing}, pages 7354--7370, Abu Dhabi, United Arab
  Emirates. Association for Computational Linguistics.

\bibitem[{Liu et~al.(2020{\natexlab{a}})Liu, Gu, Goyal, Li, Edunov,
  Ghazvininejad, Lewis, and Zettlemoyer}]{10.1162/tacl_a_00343}
Yinhan Liu, Jiatao Gu, Naman Goyal, Xian Li, Sergey Edunov, Marjan
  Ghazvininejad, Mike Lewis, and Luke Zettlemoyer. 2020{\natexlab{a}}.
\newblock \href {https://doi.org/10.1162/tacl_a_00343} {{Multilingual Denoising
  Pre-training for Neural Machine Translation}}.
\newblock \emph{Transactions of the Association for Computational Linguistics},
  8:726--742.

\bibitem[{Liu et~al.(2020{\natexlab{b}})Liu, Gu, Goyal, Li, Edunov,
  Ghazvininejad, Lewis, and Zettlemoyer}]{liu-etal-2020-multilingual-denoising}
Yinhan Liu, Jiatao Gu, Naman Goyal, Xian Li, Sergey Edunov, Marjan
  Ghazvininejad, Mike Lewis, and Luke Zettlemoyer. 2020{\natexlab{b}}.
\newblock \href {https://doi.org/10.1162/tacl_a_00343} {Multilingual denoising
  pre-training for neural machine translation}.
\newblock \emph{Transactions of the Association for Computational Linguistics},
  8:726--742.

\bibitem[{Majumder et~al.(2019)Majumder, Li, Ni, and
  McAuley}]{majumder-etal-2019-generating}
Bodhisattwa~Prasad Majumder, Shuyang Li, Jianmo Ni, and Julian McAuley. 2019.
\newblock \href {https://doi.org/10.18653/v1/D19-1613} {Generating personalized
  recipes from historical user preferences}.
\newblock In \emph{Proceedings of the 2019 Conference on Empirical Methods in
  Natural Language Processing and the 9th International Joint Conference on
  Natural Language Processing (EMNLP-IJCNLP)}, pages 5976--5982, Hong Kong,
  China. Association for Computational Linguistics.

\bibitem[{Marin et~al.(2019)Marin, Biswas, Ofli, Hynes, Salvador, Aytar, Weber,
  and Torralba}]{marin2019learning}
Javier Marin, Aritro Biswas, Ferda Ofli, Nicholas Hynes, Amaia Salvador, Yusuf
  Aytar, Ingmar Weber, and Antonio Torralba. 2019.
\newblock \href {http://pic2recipe.csail.mit.edu/tpami19.pdf} {Recipe1{M}+: A
  dataset for learning cross-modal embeddings for cooking recipes and food
  images}.
\newblock \emph{{IEEE} Trans. Pattern Anal. Mach. Intell.}

\bibitem[{Mikolov et~al.(2013{\natexlab{a}})Mikolov, Le, and Sutskever}]{bli}
Tomas Mikolov, Quoc~V. Le, and Ilya Sutskever. 2013{\natexlab{a}}.
\newblock \href {https://doi.org/10.48550/ARXIV.1309.4168} {Exploiting
  similarities among languages for machine translation}.

\bibitem[{Mikolov et~al.(2013{\natexlab{b}})Mikolov, Sutskever, Chen, Corrado,
  and Dean}]{Mikolovw2v}
Tomas Mikolov, Ilya Sutskever, Kai Chen, Greg Corrado, and Jeffrey Dean.
  2013{\natexlab{b}}.
\newblock \href
  {https://papers.nips.cc/paper/5021-distributed-representations-of-words-and-phrases-and-their-compositionality.pdf}
  {Distributed representations of words and phrases and their
  compositionality}.
\newblock In \emph{Neural and Information Processing System (NIPS)}.

\bibitem[{Morales-Garz{\'o}n et~al.(2021{\natexlab{a}})Morales-Garz{\'o}n,
  G{\'o}mez-Romero, and Martin-Bautista}]{morales-garzon-etal-2021-semantic}
Andrea Morales-Garz{\'o}n, Juan G{\'o}mez-Romero, and Maria~J. Martin-Bautista.
  2021{\natexlab{a}}.
\newblock \href {https://doi.org/10.18653/v1/2021.eacl-srw.20} {Semantic-aware
  transformation of short texts using word embeddings: An application in the
  food computing domain}.
\newblock In \emph{Proceedings of the 16th Conference of the European Chapter
  of the Association for Computational Linguistics: Student Research Workshop},
  pages 148--154, Online. Association for Computational Linguistics.

\bibitem[{Morales-Garz{\'o}n et~al.(2022)Morales-Garz{\'o}n, G{\'o}mez-Romero,
  and Mart{\'i}n-Bautista}]{10.1007/978-3-031-08974-9_24}
Andrea Morales-Garz{\'o}n, Juan G{\'o}mez-Romero, and Maria~J.
  Mart{\'i}n-Bautista. 2022.
\newblock Contextual sentence embeddings for obtaining food recipe versions.
\newblock In \emph{Information Processing and Management of Uncertainty in
  Knowledge-Based Systems}, pages 306--316, Cham. Springer International
  Publishing.

\bibitem[{Morales-Garz{\'o}n et~al.(2021{\natexlab{b}})Morales-Garz{\'o}n,
  {Gómez-Romero}, and {Martin-Bautista}}]{9351987}
Andrea Morales-Garz{\'o}n, J.~{Gómez-Romero}, and M.~J. {Martin-Bautista}.
  2021{\natexlab{b}}.
\newblock \href {https://doi.org/10.1109/ACCESS.2021.3058559} {A word
  embedding-based method for unsupervised adaptation of cooking recipes}.
\newblock \emph{IEEE Access}, pages 1--1.

\bibitem[{Mori et~al.(2014)Mori, Maeta, Yamakata, and Sasada}]{mori2014flow}
Shinsuke Mori, Hirokuni Maeta, Yoko Yamakata, and Tetsuro Sasada. 2014.
\newblock \href
  {http://www.lrec-conf.org/proceedings/lrec2014/pdf/763_Paper.pdf} {Flow graph
  corpus from recipe texts}.
\newblock In \emph{Proceedings of the Ninth International Conference on
  Language Resources and Evaluation ({LREC}'14)}, pages 2370--2377, Reykjavik,
  Iceland. European Language Resources Association (ELRA).

\bibitem[{Muennighoff et~al.(2022)Muennighoff, Wang, Sutawika, Roberts,
  Biderman, Scao, Bari, Shen, Yong, Schoelkopf
  et~al.}]{muennighoff2022crosslingual}
Niklas Muennighoff, Thomas Wang, Lintang Sutawika, Adam Roberts, Stella
  Biderman, Teven~Le Scao, M~Saiful Bari, Sheng Shen, Zheng-Xin Yong, Hailey
  Schoelkopf, et~al. 2022.
\newblock Crosslingual generalization through multitask finetuning.
\newblock \emph{arXiv preprint arXiv:2211.01786}.

\bibitem[{Naous et~al.(2023)Naous, Ryan, and Xu}]{naous2023having}
Tarek Naous, Michael~J Ryan, and Wei Xu. 2023.
\newblock Having beer after prayer? measuring cultural bias in large language
  models.
\newblock \emph{arXiv preprint arXiv:2305.14456}.

\bibitem[{OpenAI(2023)}]{openai2023gpt4}
OpenAI. 2023.
\newblock \href {http://arxiv.org/abs/2303.08774} {{GPT}-4 technical report}.

\bibitem[{Ouyang et~al.(2022)Ouyang, Wu, Jiang, Almeida, Wainwright, Mishkin,
  Zhang, Agarwal, Slama, Ray et~al.}]{ouyang2022training}
Long Ouyang, Jeffrey Wu, Xu~Jiang, Diogo Almeida, Carroll Wainwright, Pamela
  Mishkin, Chong Zhang, Sandhini Agarwal, Katarina Slama, Alex Ray, et~al.
  2022.
\newblock Training language models to follow instructions with human feedback.
\newblock \emph{Advances in Neural Information Processing Systems},
  35:27730--27744.

\bibitem[{Palta and Rudinger(2023)}]{palta-rudinger-2023-fork}
Shramay Palta and Rachel Rudinger. 2023.
\newblock \href {https://aclanthology.org/2023.findings-acl.631} {{FORK}: A
  bite-sized test set for probing culinary cultural biases in commonsense
  reasoning models}.
\newblock In \emph{Findings of the Association for Computational Linguistics:
  ACL 2023}, pages 9952--9962, Toronto, Canada. Association for Computational
  Linguistics.

\bibitem[{Papineni et~al.(2002)Papineni, Roukos, Ward, and
  Zhu}]{papineni-etal-2002-bleu}
Kishore Papineni, Salim Roukos, Todd Ward, and Wei-Jing Zhu. 2002.
\newblock \href {https://doi.org/10.3115/1073083.1073135} {{B}leu: a method for
  automatic evaluation of machine translation}.
\newblock In \emph{Proceedings of the 40th Annual Meeting of the Association
  for Computational Linguistics}, pages 311--318, Philadelphia, Pennsylvania,
  USA. Association for Computational Linguistics.

\bibitem[{Peskov et~al.(2021)Peskov, Hangya, Boyd-Graber, and
  Fraser}]{peskov-etal-2021-adapting-entities}
Denis Peskov, Viktor Hangya, Jordan Boyd-Graber, and Alexander Fraser. 2021.
\newblock \href {https://doi.org/10.18653/v1/2021.findings-emnlp.315} {Adapting
  entities across languages and cultures}.
\newblock In \emph{Findings of the Association for Computational Linguistics:
  EMNLP 2021}, pages 3725--3750, Punta Cana, Dominican Republic. Association
  for Computational Linguistics.

\bibitem[{Piktus et~al.(2023)Piktus, Akiki, Villegas, Lauren{\c{c}}on, Dupont,
  Luccioni, Jernite, and Rogers}]{piktus-etal-2023-roots}
Aleksandra Piktus, Christopher Akiki, Paulo Villegas, Hugo Lauren{\c{c}}on,
  G{\'e}rard Dupont, Sasha Luccioni, Yacine Jernite, and Anna Rogers. 2023.
\newblock \href {https://aclanthology.org/2023.acl-demo.29} {The {ROOTS} search
  tool: Data transparency for {LLM}s}.
\newblock In \emph{Proceedings of the 61st Annual Meeting of the Association
  for Computational Linguistics (Volume 3: System Demonstrations)}, pages
  304--314, Toronto, Canada. Association for Computational Linguistics.

\bibitem[{Popovi{\'c}(2015)}]{popovic-2015-chrf}
Maja Popovi{\'c}. 2015.
\newblock \href {https://doi.org/10.18653/v1/W15-3049} {chr{F}: character
  n-gram {F}-score for automatic {MT} evaluation}.
\newblock In \emph{Proceedings of the Tenth Workshop on Statistical Machine
  Translation}, pages 392--395, Lisbon, Portugal. Association for Computational
  Linguistics.

\bibitem[{Post(2018)}]{post-2018-call}
Matt Post. 2018.
\newblock \href {https://doi.org/10.18653/v1/W18-6319} {A call for clarity in
  reporting {BLEU} scores}.
\newblock In \emph{Proceedings of the Third Conference on Machine Translation:
  Research Papers}, pages 186--191, Brussels, Belgium. Association for
  Computational Linguistics.

\bibitem[{Radford et~al.(2019)Radford, Wu, Child, Luan, Amodei, and
  Sutskever}]{radford2019language}
Alec Radford, Jeff Wu, Rewon Child, David Luan, Dario Amodei, and Ilya
  Sutskever. 2019.
\newblock \href
  {https://d4mucfpksywv.cloudfront.net/better-language-models/language-models.pdf}
  {Language models are unsupervised multitask learners}.

\bibitem[{Rebechi and da~Silva(2017)}]{10.1007/978-3-319-69805-2_8}
Rozane~Rodrigues Rebechi and M{\'a}rcia~Moura da~Silva. 2017.
\newblock Brazilian recipes in {P}ortuguese and {E}nglish: The role of
  phraseology for translation.
\newblock In \emph{Computational and Corpus-Based Phraseology}, pages 102--114,
  Cham. Springer International Publishing.

\bibitem[{Reiter(2018)}]{reiter-2018-structured}
Ehud Reiter. 2018.
\newblock \href {https://doi.org/10.1162/coli_a_00322} {A structured review of
  the validity of {BLEU}}.
\newblock \emph{Computational Linguistics}, 44(3):393--401.

\bibitem[{Salvador et~al.(2017)Salvador, Hynes, Aytar, Marin, Ofli, Weber, and
  Torralba}]{salvador2017learning}
Amaia Salvador, Nicholas Hynes, Yusuf Aytar, Javier Marin, Ferda Ofli, Ingmar
  Weber, and Antonio Torralba. 2017.
\newblock Learning cross-modal embeddings for cooking recipes and food images.
\newblock In \emph{Proceedings of the IEEE conference on computer vision and
  pattern recognition}, pages 3020--3028.

\bibitem[{Scao et~al.(2022)Scao, Fan, Akiki, Pavlick, Ili{\'c}, Hesslow,
  Castagn{\'e}, Luccioni, Yvon, Gall{\'e} et~al.}]{scao2022bloom}
Teven~Le Scao, Angela Fan, Christopher Akiki, Ellie Pavlick, Suzana Ili{\'c},
  Daniel Hesslow, Roman Castagn{\'e}, Alexandra~Sasha Luccioni, Fran{\c{c}}ois
  Yvon, Matthias Gall{\'e}, et~al. 2022.
\newblock {BLOOM}: A 176b-parameter open-access multilingual language model.
\newblock \emph{arXiv preprint arXiv:2211.05100}.

\bibitem[{S{\o}gaard et~al.(2018)S{\o}gaard, Ruder, and
  Vuli{\'c}}]{sogaard-etal-2018-limitations}
Anders S{\o}gaard, Sebastian Ruder, and Ivan Vuli{\'c}. 2018.
\newblock \href {https://doi.org/10.18653/v1/P18-1072} {On the limitations of
  unsupervised bilingual dictionary induction}.
\newblock In \emph{Proceedings of the 56th Annual Meeting of the Association
  for Computational Linguistics (Volume 1: Long Papers)}, pages 778--788,
  Melbourne, Australia. Association for Computational Linguistics.

\bibitem[{Song et~al.(2020)Song, Tan, Qin, Lu, and Liu}]{NEURIPS2020_c3a690be}
Kaitao Song, Xu~Tan, Tao Qin, Jianfeng Lu, and Tie-Yan Liu. 2020.
\newblock \href
  {https://proceedings.neurips.cc/paper/2020/file/c3a690be93aa602ee2dc0ccab5b7b67e-Paper.pdf}
  {Mpnet: Masked and permuted pre-training for language understanding}.
\newblock In \emph{Advances in Neural Information Processing Systems},
  volume~33, pages 16857--16867. Curran Associates, Inc.

\bibitem[{Tang et~al.(2020)Tang, Tran, Li, Chen, Goyal, Chaudhary, Gu, and
  Fan}]{https://doi.org/10.48550/arxiv.2008.00401}
Yuqing Tang, Chau Tran, Xian Li, Peng-Jen Chen, Naman Goyal, Vishrav Chaudhary,
  Jiatao Gu, and Angela Fan. 2020.
\newblock \href {https://doi.org/10.48550/ARXIV.2008.00401} {Multilingual
  translation with extensible multilingual pretraining and finetuning}.

\bibitem[{Tiedemann and Thottingal(2020)}]{TiedemannThottingal:EAMT2020}
J{\"o}rg Tiedemann and Santhosh Thottingal. 2020.
\newblock {OPUS-MT} — {B}uilding open translation services for the {W}orld.
\newblock In \emph{Proceedings of the 22nd Annual Conferenec of the European
  Association for Machine Translation (EAMT)}, Lisbon, Portugal.

\bibitem[{Wang et~al.(2021)Wang, Tu, Tan, Wang, Sun, and
  Liu}]{https://doi.org/10.48550/arxiv.2106.13627}
Shuo Wang, Zhaopeng Tu, Zhixing Tan, Wenxuan Wang, Maosong Sun, and Yang Liu.
  2021.
\newblock \href {https://doi.org/10.48550/ARXIV.2106.13627} {Language models
  are good translators}.

\bibitem[{Wein et~al.(2022)Wein, Leung, Mu, and
  Schneider}]{wein-etal-2022-effect}
Shira Wein, Wai~Ching Leung, Yifu Mu, and Nathan Schneider. 2022.
\newblock \href {https://aclanthology.org/2022.law-1.12} {Effect of source
  language on {AMR} structure}.
\newblock In \emph{Proceedings of the 16th Linguistic Annotation Workshop
  (LAW-XVI) within LREC2022}, pages 97--102, Marseille, France. European
  Language Resources Association.

\bibitem[{Xue et~al.(2021)Xue, Constant, Roberts, Kale, Al-Rfou, Siddhant,
  Barua, and Raffel}]{xue-etal-2021-mt5}
Linting Xue, Noah Constant, Adam Roberts, Mihir Kale, Rami Al-Rfou, Aditya
  Siddhant, Aditya Barua, and Colin Raffel. 2021.
\newblock \href {https://doi.org/10.18653/v1/2021.naacl-main.41} {m{T}5: A
  massively multilingual pre-trained text-to-text transformer}.
\newblock In \emph{Proceedings of the 2021 Conference of the North American
  Chapter of the Association for Computational Linguistics: Human Language
  Technologies}, pages 483--498, Online. Association for Computational
  Linguistics.

\bibitem[{Yamakata et~al.(2017)Yamakata, Carroll, and
  Mori}]{yamakata2017comparison}
Yoko Yamakata, John Carroll, and Shinsuke Mori. 2017.
\newblock A comparison of cooking recipe named entities between {J}apanese and
  {E}nglish.
\newblock In \emph{Proceedings of the 9th Workshop on Multimedia for Cooking
  and Eating Activities in conjunction with The 2017 International Joint
  Conference on Artificial Intelligence}, pages 7--12.

\bibitem[{Yamakata et~al.(2016)Yamakata, Imahori, Maeta, and
  Mori}]{yamakata2016method}
Yoko Yamakata, Shinji Imahori, Hirokuni Maeta, and Shinsuke Mori. 2016.
\newblock \href {https://doi.org/10.1109/ICMEW.2016.7574705} {A method for
  extracting major workflow composed of ingredients, tools, and actions from
  cooking procedural text}.
\newblock In \emph{2016 IEEE International Conference on Multimedia \& Expo
  Workshops (ICMEW)}.

\bibitem[{Zeng et~al.(2022)Zeng, Liu, Du, Wang, Lai, Ding, Yang, Xu, Zheng, Xia
  et~al.}]{zeng2022glm}
Aohan Zeng, Xiao Liu, Zhengxiao Du, Zihan Wang, Hanyu Lai, Ming Ding, Zhuoyi
  Yang, Yifan Xu, Wendi Zheng, Xiao Xia, et~al. 2022.
\newblock Glm-130b: An open bilingual pre-trained model.
\newblock \emph{arXiv preprint arXiv:2210.02414}.

\bibitem[{Zhang et~al.(2019{\natexlab{a}})Zhang, Trattner, Ludwig, and
  Elsweiler}]{zhang2019understanding}
Qing Zhang, Christoph Trattner, Bernd Ludwig, and David Elsweiler.
  2019{\natexlab{a}}.
\newblock Understanding cross-cultural visual food tastes with online recipe
  platforms.
\newblock In \emph{Proceedings of the International AAAI Conference on Web and
  Social Media}, volume~13, pages 671--674.

\bibitem[{Zhang et~al.(2019{\natexlab{b}})Zhang, Kishore, Wu, Weinberger, and
  Artzi}]{zhang2019bertscore}
Tianyi Zhang, Varsha Kishore, Felix Wu, Kilian~Q Weinberger, and Yoav Artzi.
  2019{\natexlab{b}}.
\newblock \href {https://openreview.net/forum?id=SkeHuCVFDr} {{BERTScore}:
  Evaluating text generation with {BERT}}.
\newblock In \emph{International Conference on Learning Representations}.

\bibitem[{Zhou et~al.(2023{\natexlab{a}})Zhou, Cabello, Cao, and
  Hershcovich}]{zhou-etal-2023-cross}
Li~Zhou, Laura Cabello, Yong Cao, and Daniel Hershcovich. 2023{\natexlab{a}}.
\newblock \href {https://aclanthology.org/2023.c3nlp-1.2} {Cross-cultural
  transfer learning for {C}hinese offensive language detection}.
\newblock In \emph{Proceedings of the First Workshop on Cross-Cultural
  Considerations in NLP (C3NLP)}, pages 8--15, Dubrovnik, Croatia. Association
  for Computational Linguistics.

\bibitem[{Zhou et~al.(2023{\natexlab{b}})Zhou, Karamolegkou, Chen, and
  Hershcovich}]{zhou2023cultural}
Li~Zhou, Antonia Karamolegkou, Wenyu Chen, and Daniel Hershcovich.
  2023{\natexlab{b}}.
\newblock Cultural compass: Predicting transfer learning success in offensive
  language detection with cultural features.
\newblock \emph{arXiv preprint arXiv:2310.06458}.

\end{thebibliography}
\bibliographystyle{acl_natbib}

\end{document}